\documentclass[authoryear, preprint,review,12pt]{elsarticle}

\usepackage{graphicx}
\usepackage{amssymb}
\usepackage{subfigure}
\usepackage{multirow}
\usepackage{lineno}
\usepackage{float}
\usepackage{epstopdf}
\usepackage{siunitx}
\usepackage{array}
\usepackage{mathtools}
\usepackage{amsmath}
\usepackage{color}
\usepackage{nomencl}

\usepackage{natbib}
\setcitestyle{citesep={;}, aysep={,}}

\makenomenclature

\journal{Biosystems Engineering}

\begin{document}

\begin{frontmatter}

\title{Terrain assessment for precision agriculture using vehicle dynamic modelling}
\author[label1]{Giulio Reina}
\author[label2]{Annalisa Milella}
\author[label1]{Rocco Galati}

\address[label1]{Department of Engineering for Innovation, University of Salento, Via Arnesano, 73100 Lecce, Italy, Corresponding author Email: giulio.reina@unisalento.it}
\address[label2]{Institute of Intelligent Systems for Automation, CNR, Via Amendola 122 D/O, 70126 Bari, Italy}

\begin{abstract}
Advances in precision agriculture greatly rely on innovative control and sensing technologies that allow service units to increase their level of driving automation while ensuring at the same time high safety standards. This paper deals with automatic terrain estimation and classification that is performed simultaneously by an agricultural vehicle during normal operations. Vehicle mobility and safety, and the successful implementation of important agricultural tasks including seeding, plowing, fertilising and controlled traffic depend or can be improved by a correct identification of the terrain that is traversed. The novelty of this research lies in that terrain estimation is performed by using not only traditional appearance-based features, that is colour and geometric properties, but also contact-based features, that is measuring physics-based dynamic effects that govern the vehicle-terrain interaction and that greatly affect its mobility. Experimental results obtained from an all-terrain vehicle operating on different surfaces are presented to validate the system in the field. It was shown that a terrain classifier trained with contact features was able to achieve a correct prediction rate of 85.1\%, which is comparable or better than that obtained with approaches using traditional feature sets. To further improve the classification performance, all feature sets were merged in an augmented feature space, reaching, for these tests, 89.1\% of correct predictions.
\end{abstract}

\begin{keyword}

Self-driving vehicles \sep vehicle dynamics \sep agricultural robotics  \sep terrain estimation and classification\sep
  vehicle-terrain interaction \sep sensor data processing \sep visual and range sensing
\end{keyword}

\end{frontmatter}

\mbox{}

\nomenclature{RGB}{Red Green Blue colour space}
\nomenclature{r$_p$, g$_p$, b$_p$}{Red, Green, and Blue channel}
\nomenclature{RGB-D}{Red Green Blue-Depth}
\nomenclature{VO}{Visual Odometry}
\nomenclature{$B, L$}{Vehicle width and length, $m$}
\nomenclature{$W$}{Vehicle weight, $N$}
\nomenclature{$k_t$}{Motor torque constant, $NmA^{-1}$}
\nomenclature{$\tau$}{Motor gearhead ratio}
\nomenclature{SVM}{Support Vector Machine}
\nomenclature{4WD}{Four wheel drive}
\nomenclature{IMU}{Inertial measurement unit (IMU)}
\nomenclature{SVD}{Singular value decomposition}
\nomenclature{WRF, VRF}{World and Vehicle Reference Frame}
\nomenclature{RPY}{Roll, Pitch and Yaw angle convention}
\nomenclature{QV}{Quarter-vehicle model}
\nomenclature{$\theta_p$}{Average slope of a terrain patch, $rad$}
\nomenclature{$\sigma_z^{2}$}{Height variance, $m^{2}$}
\nomenclature{$c_1c_2c_3$}{The $c_1c_2c_3$ colour space}
\nomenclature{$E_i$}{Mean pixel intensity value}
\nomenclature{$\sigma_i^{2}$}{Variance of pixel intensity value}
\nomenclature{$SK_i, Ku_i$}{Skewness and Kurtosis of pixel intensity value}
\nomenclature{$x_n$}{Pixel intensity value}
\nomenclature{$N$}{Total number of points in a terrain patch}
\nomenclature{$\hat{n}_p$}{Normal unit vector of a terrain patch}
\nomenclature{$\hat{z}$}{Z-axis unit vector}
\nomenclature{$F$}{Goodness of fit}
\nomenclature{$\lambda_i$}{Minimum singular value of points' covariance matrix}
\nomenclature{$C_p$}{Points' covariance matrix}
\nomenclature{$\Delta{z}$}{Height range, $m$}
\nomenclature{$z_{max}, z_{min}$}{Maximum and minimum point coordinate value along Z-axis, $m$}
\nomenclature{$\omega$}{Tyre angular velocity, $rads^{-1}$}
\nomenclature{$\omega_e$}{Excitation frequency, $Hz$}
\nomenclature{$\lambda$}{Irregularity wavelength, $m$}
\nomenclature{$W_v$}{Tyre vertical load, $N$}
\nomenclature{$s_z$}{Normal stress, $Nm^{-2}$}
\nomenclature{$T_r$}{Driving torque, $Nm$}
\nomenclature{$r$}{Wheel radius, $m$}
\nomenclature{$f_r$}{Coefficient of motion resistance}
\nomenclature{$I$}{Electrical current, $A$}
\nomenclature{$\theta$, $\phi$, $\psi$}{Vehicle pitch, roll, and yaw angle, $rad$}
\nomenclature{$R^{W}_{V}$}{Rotation matrix from vehicle to world reference frame}
\nomenclature{$F^{W}, F^{V}$}{Vehicle weight force in the world and vehicle reference frame, $N$}
\nomenclature{$h$}{Height of the center of gravity above the ground, $m$}
\nomenclature{$F_{z,i}$}{Vertical force on wheel $i$, $N$}
\nomenclature{$S$}{Slip}
\nomenclature{$V$}{Actual forward vehicle speed, $ms^{-1}$}
\nomenclature{$m_b$}{Sprung mass, $kg$}
\nomenclature{$z_b$}{Sprung mass vertical displacement, $m$}
\nomenclature{$k_w$}{Vertical wheel stiffness coefficient, $Nm^{-1}$}
\nomenclature{$c_w$}{Vertical wheel damping coefficient, $Nsm^{-1}$}
\nomenclature{$z_d$}{Vertical displacement of terrain profile, $m$}
\nomenclature{$H_a$}{Magnitude of transfer function for acceleration, $s^{-2}$}

\printnomenclature

\section{Introduction}
Latest research efforts in robotic mobility are devoted to the development of novel perception systems that allow vehicles to travel long distances with limited or no human supervision in difficult
scenarios, including planetary exploration, mining, all-terrain self-driving cars, and agricultural vehicles. One of the challenges in agricultural robotics is the ability to perceive and analyse the traversed ground.  The knowledge of the type of terrain can be beneficial for a vehicle to deal with its environment more efficiently and to better support precision farming tasks. It is known that on natural terrain, wheel-terrain interaction has a critical influence on vehicle mobility that can be very different on ploughed terrain rather than on dirt road or compacted soil. Therefore, locomotion performance can be optimised in terms of traction or power consumption (e.g., fuel or battery life) by adapting control and planning strategy to site-specific environments.\\ Terrain estimation would also contribute to increase the safety of agricultural vehicles during operations near ditches or on hillsides and cross slopes and on hazardous highly-deformable terrain. Another important aspect that is raising interest in precision agriculture is related with the prediction of the risk of soil compaction by farm machinery that can be drawn from monitoring terrain parameters related with the ability to support vehicular motion \citep{STE}.

This paper addresses the problem of terrain assessment for highly-automated vehicles. This issue can be divided into two sub-problems: terrain characterisation and classification. Terrain characterisation deals with the determination of distinctive traits or features that well describe and identify a certain class of terrain. Based on these features, terrain can be classified by association with one of the predefined, commonly known, categories, such as ploughed terrain, dirt road, etc.\\
In this research, it is shown that terrain identification can be performed based on geometric and visual appearance of the terrain, that is using common exteroceptive features, as well as based on measures pertaining to vehicle-terrain interaction, that is resorting to proprioceptive or contact features. Overall, three sets are considered, namely colour, geometric and contact features. The colour set accounts for the normalised RGB content of a given terrain. The geometric block refers to statistics obtained from 3D point coordinates associated with terrain patches. Finally, the contact block describes the vehicle dynamic response to a given terrain in terms of wheel slip, rolling resistance and body vibration.\\
Using a supervised classifier, the association of a given terrain under inspection with a few predefined agricultural surfaces is investigated. The colour, geometric and contact feature sets are used first singularly and, then, in combination, showing their advantages and disadvantages for terrain classification.\\ Experimental results are included to validate the proposed approach using an all-terrain vehicle operating in agricultural settings.\\     The research is presented in the paper as follows. Section \ref{s2} surveys related research in the literature, pointing out the differences and novel contributions of this work. Section 3 provides an overview of the framework proposed for terrain estimation and classification, whereas a description of the test bed used for the testing and development of the system is presented in Section 4. Terrain estimation through feature selection is thoroughly discussed in Section 5, along with practical examples. In Section 6, extensive experimental results obtained from field tests are included to support and evaluate quantitatively the proposed terrain classier. Final remarks are drawn in Section 7.

\section{Literature review}\label{s2}

Previous work on terrain estimation in robotic mobility has mostly relied on the use of colour and geometric features that are extracted from sensory data acquired by exteroceptive sensors, including vision \citep{ROS, REI2}, radar \citep*{REIJ}, and lidar \citep*{KRA}. Agricultural soil was further analysed with a 3D laser scanner in \cite{FER}, based on three main geometric parameters: root mean square of the height variations, autocorrelation function and correlation length. In addition, range and intensity measurements obtained from a ground lidar were used to discriminate between ground, weed and maize \citep{AND}.\\ Alternatively, hyperspectral imaging was used to identify and classify terrain \citep{LU}. Approaches based on RGB-D cameras have also been used for terrain estimation. For example, the Kinect sensor was used in \cite{MA} to characterise soil during tillage operations, and to develop a navigation system for greenhouse operations \citep{NI}. In all these examples, depth and visual information (i.e., range, colour, and texture) were employed.\\
Recent research has also focused on terrain estimation via proprioceptive sensors or proprioceptors. For example, a method to estimate terrain cohesion and internal friction angle was proposed using ``internal" sensors and terramechanics theory \citep*{IAG}. Motor currents and rate-of-turn of a robot were also correlated with soil parameters \citep*{OJE}. Vibrations induced by wheel-ground interaction were fed to different types of terrain classifiers that discriminate based on, respectively, Mahalanobis distance of the power spectral densities \citep{IAGV}, AdaBoost \citep*{KRE}, and neural network \citep*{DUP}. Finally, a method for terrain characterisation was proposed using measures of slippage incurred by a skid-steer robot during turning motion \citep{REIV}.\\ From this survey, it appears that research has focused either on appearance-based features or on parameters pertaining to wheel-terrain interaction. Little attention has been devoted to the combination of these two aspects. For example, exteroceptive and proprioceptive features were applied separately showing comparable results but no attempt was made to combine them \citep*{KRE}. A self-supervised learning framework was proposed using a proprioceptive terrain classifier to train an exteroceptive (i.e., vision-based) terrain classifier \citep{Brooks2007}.\\
In this paper, terrain estimation and classification is tackled by combining exteroceptive with proprioceptive features. The ability of each class in discriminating terrain is investigated singularly and using different feature combinations. The main contribution of this research is the adoption of a multimodal description of the terrain that combines contactless sensing with contact measurements. One obvious advantage is that terrain classification gains in robustness and it can be performed even when some of the feature sets are not usable. For example, visual estimates may be corrupted due to low-lighting conditions or heavy shadowing. This rich source of information may be made available to the user and presented within a multimodal interactive map or farm management information system, using standard tablet PCs, where he/she can browse between different information layers, i.e., RGB map, digital elevation map or traversability and trafficability map.\\In addition, learning the relationship between the visual appearance of terrain and its influence on vehicle mobility, would allow to identify potentially hazardous surfaces from a distance by extending the near-range  classification results to far-range via segmentation of the entire visual image.

\section{Overview of proposed approach}
\begin{figure}[t]
      \centering
      \includegraphics[width=12.0 cm]{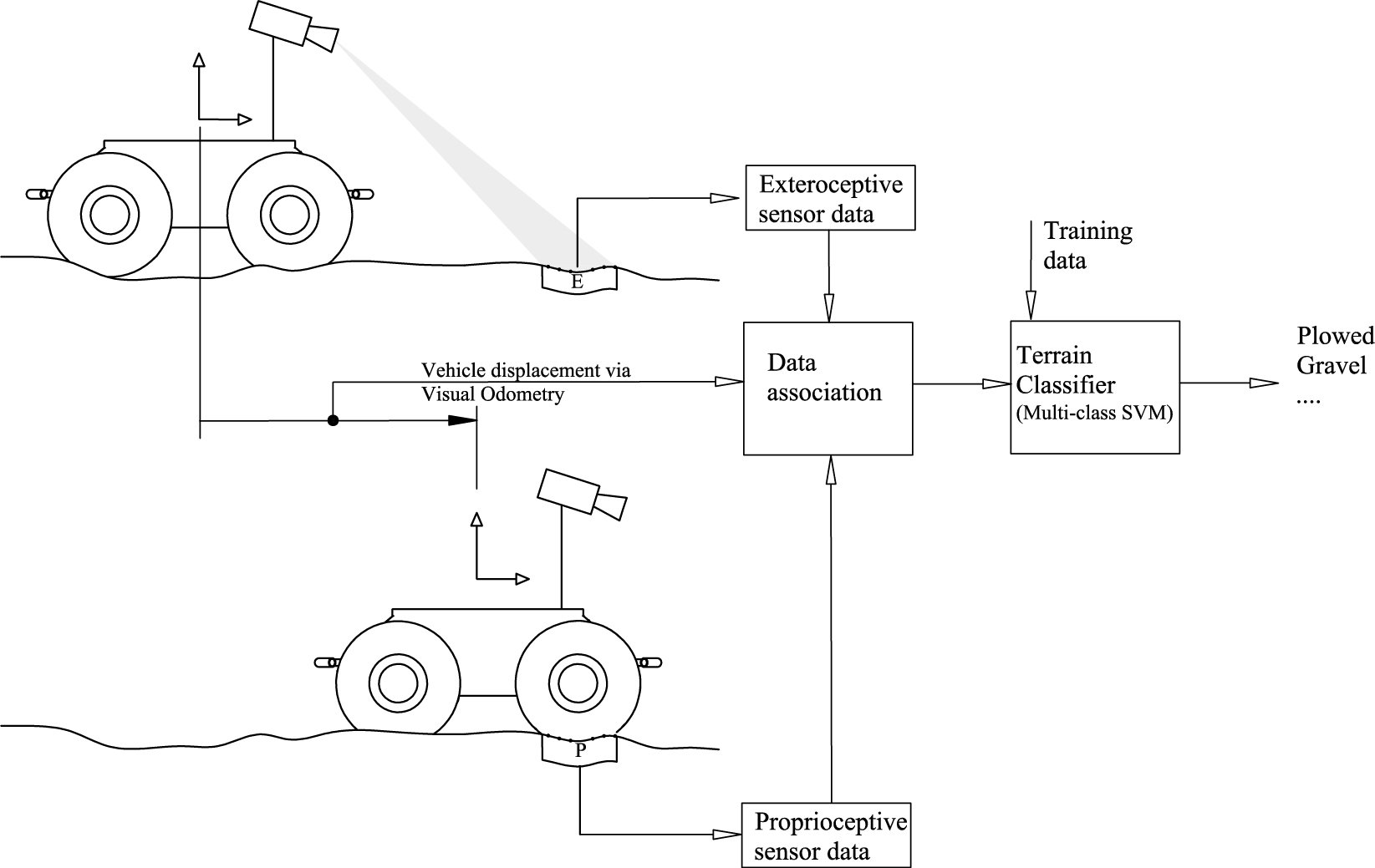}
      \caption{Pipeline of the supervised terrain classifier using exteroceptive and proprioceptive data}
      \label{f1}
\end{figure}
A supervised learning method is proposed to recognise terrain surface for a vehicle operating in rural settings. The algorithm learns to identify distinct terrain types based on labelled proprioceptive and exteroceptive data provided in a training phase. During training, the algorithm analyses these data sets to form a model of the signals corresponding to each terrain class. Once trained, the classifier can be applied on-line during normal operations and new observations can be automatically associated with one of the labelled terrain classes. A schematic of the algorithm is shown in Figure \ref{f1}.\\
One important aspect of this research is that visual sensors are generally mounted in a forward-looking configuration. At a given time instant, visual and proprioceptive data measure different portions of the environment, that is, the stereocamera surveys the environment in front of the vehicle, whereas proprioceptors sense the current supporting surface. Therefore, classification follows a two-step process, as explained in Figure \ref{f1}. First, a given terrain patch is characterised from a distance and a vision-based feature set, E, can be assigned to it. Then, when the same patch is traversed by the vehicle, a proprioceptive feature set, P, can also be attached, and the patch can be classified. The match of the exteroceptive and proprioceptive feature set requires the estimation of the successive displacements of the vehicle via a visual odometry (VO) algorithm. Alternatively, some global position estimation could also be employed. However, in this work, VO is the preferred approach, as it provides a convenient self-contained localisation system. \\An error-correcting output code multiclass model \citep{FUR} is employed to classify the terrain into multiple categories. This approach reduces the problem of classification with three or more classes to a set of binary classifiers. Support vector machines (SVM) learners with linear kernel were employed, with a one-vs-one coding design (i.e., for each binary learner, one class is set as positive, another is set as negative, and the rest are ignored, exhausting all combinations of positive class assignments). \\As a positive byproduct of the classification process, by stitching together portions of the environment progressively traversed by the vehicle, it is possible to build a map that includes 3D data and RGB content (exteroceptive data) with motion resistance, slip and vibrational response (proprioceptive data) experienced by the vehicle.
\section{System architecture}\label{s4}

\begin{figure}[]
      \centering
      \includegraphics[width=8.0 cm]{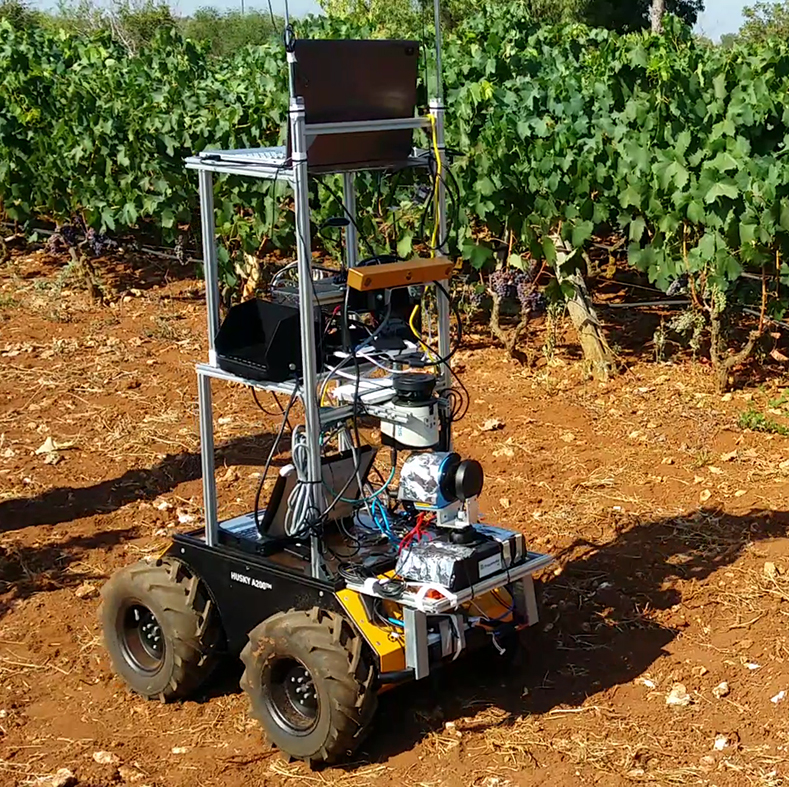}
      \caption{The all-terrain vehicle used for field validation}
      \label{f2}
\end{figure}

The Husky A200 robotic platform was used for experimental sensor data gathering. The vehicle, shown in Figure \ref{f2}, is a non-holonomic four-wheel drive (4WD) skid-steer robot, whose salient technical data are listed in Table \ref{t1}. The proprioceptive sensor suite includes electrical current and voltage sensors to measure wheel mechanical torque, encoders to estimate wheel angular velocity, and an inertial measurement unit (IMU) (XSENS MTi-10) that outputs the vehicle angular rate and acceleration measurements, and estimated navigation body attitude (i.e., Euler angles: roll, pitch, yaw). Exteroceptive perception is performed via a colour stereo camera (Point Grey XB3) that is mounted on a dedicated aluminum frame. It provides three-dimensional reconstruction with RGB colour data of the environment in front of the vehicle up to a look-ahead distance of about 30 m with a 0.092 m average accuracy and a 0.062 m standard deviation \citep*{reina2016}.

\begin{table}[]
\begin{center}
{\normalsize
\begin{tabular}{l l}
\hline \hline Dimensions (Width $\times$ Length)& B$\times$L=0.54 m$\times$0.7 m\\
\hline \hline Vehicle type & 4WD\\
\hline \hline Tyre size & 13$\times$5-6 NHS\\
 \hline Total weight (sensor payload included) & W=313.6 N\\
\hline Motor torque constant & $k_t$=0.044 Nm/A\\
\hline Motor Gearhead ratio & $\tau=$78.71 : 1\\
\hline
\hline
\end{tabular}
}
\end{center}
 \caption{Specifications of the all-terrain vehicle Husky}
\label{t1}
\end{table}

\section{Terrain characterisation}
In this section, the feature set adopted for terrain estimation is presented. First, a description of the exteroceptive features is provided, then the proprioceptive counterparts are detailed. Practical issues connected with their estimation in the field are also addressed.
\subsection{Exteroceptive features}
\subsubsection{Colour features}
Stereocamera provides raw colour data as red, green, and blue (RGB) intensities. However, this representation suffers from large sensitivity to changes in lighting conditions, possibly leading to poor classification results. To overcome this issue, the so-called $c_1c_2c_3$ colour model can be used \citep{GEV}, i.e.,
\begin{eqnarray}\label{e_colorspace}
c_1=\arctan\left(\frac{r_p}{\max(g_p, b_p)}\right)\\
c_2=\arctan\left(\frac{g_p}{\max(r_p, b_p)}\right)\\
c_3=\arctan\left(\frac{b_p}{\max(r_p, g_p)}\right)
\end{eqnarray}
where r$_p$, g$_p$, and b$_p$ are the pixel values in the RGB space. \\For each terrain patch, colour features were obtained by using statistical moments extracted from each channel $i$ of the $c_1c_2c_3$ space as:
\begin{equation}\label{e_colorfeatures1}
E_i = \frac{1}{N}\sum_{n=1}^{N}x_n
\end{equation}
\begin{equation}\label{e_colorfeatures2}
\sigma_i^2= \frac{1}{N}\sum_{n=1}^{N}(x_n-E_i)^2
\end{equation}
\begin{equation}\label{e_colorfeatures3}
Sk_i^3 = \frac{1}{N}\sum_{n=1}^{N}(x_n-E_i)^3
\end{equation}
\begin{equation}\label{e_colorfeatures4}
Ku_i^4 = \frac{1}{N}\sum_{n=1}^{N}(x_n-E_i)^4
\end{equation}
where $x_n$ is the intensity level associated to a point in one of the three colour channels, and $N$ is the total number of points in the patch. The first moment, i.e., the mean (Eq. \ref{e_colorfeatures1}) defines where the individual colour lies in the $c_1c_2c_3$ colour space. The second moment, i.e., the variance (Eq. \ref{e_colorfeatures2}), indicates the spread or scale of the colour distribution. The third moment, i.e., the skewness (Eq. \ref{e_colorfeatures3}) provides a measure of the asymmetry of the data around the sample mean and indicates if the $c_1c_2c_3$ values lie towards maximum or minimum in the scale. The fourth moment, i.e., the Kurtosis (Eq. \ref{e_colorfeatures4}) measures the flatness or peakedness of the colour distribution. Thus, the colour properties were represented by four features for each channel for a total of twelve elements.

\subsubsection{Geometric features}
Geometric features are statistics obtained from the coordinates of the $N$ points that belong to a certain region of interest or terrain patch. The first element of the geometric feature vector was defined as the average slope of the terrain patch, that is, the angle $\theta_p$ between the least-squares-fit plane represented by its normal unit vector, $\hat{n}_p$, and the horizontal plane expressed by the $Z$-axis unit vector, $\hat{z}$. The goodness of fit, $F$, obtained as the mean-squared deviation of the points from the least-squares plane along its normal is the second feature component. Note that this corresponds to the minimum singular value, $\lambda_i$, of the points' covariance matrix, $C_p$, that is obtained using singular value decomposition (SVD). The third component is the variance in the $z$-coordinate of the patch points, $\sigma^{2}_{z}$. Finally, the fourth element is the range of point heights, ${\Delta{z}}$, defined as the difference between the maximum, $z_{max}$, and the minimum, $z_{min}$, point coordinate value along the $Z$-axis. Therefore, the geometric signature of a certain terrain patch was expressed by a four-element vector given by:
\begin{equation}
\begin{matrix*}[l]
\textrm{Slope:} & \theta_p = \arccos (\hat{n}_p \cdot \hat{z})\\
\textrm{Goodness of fit :} & F = \min (\lambda_i(C_p))\\
\textrm{Variance $z$:} & \sigma_z^2 = \frac{1}{N}\sum_{i=1}^{N} (z_i - \overline{z}_p)^2 \\
\textrm{Height range:} & \Delta{z} = |z_{max}-z_{min}|\\
\end{matrix*}
\label{eq:terFeatures}
\end{equation}
\subsection{Proprioceptive features}
Characteristic traits of the terrain can be extracted using vehicle wheels as ``tactile" sensors that generate signals modulated by the physical vehicle-terrain interaction. Being not affected by lighting conditions, this feature set can be a potentially attractive complement to vision-based features. They can also be useful for classification when a terrain is covered by a thin layer of leaves or dirt, and, in general,
when the vision sensor is occluded.
\subsubsection{Coefficient of motion resistance}
\begin{figure}[t]
      \centering
      \includegraphics[width=6.0 cm]{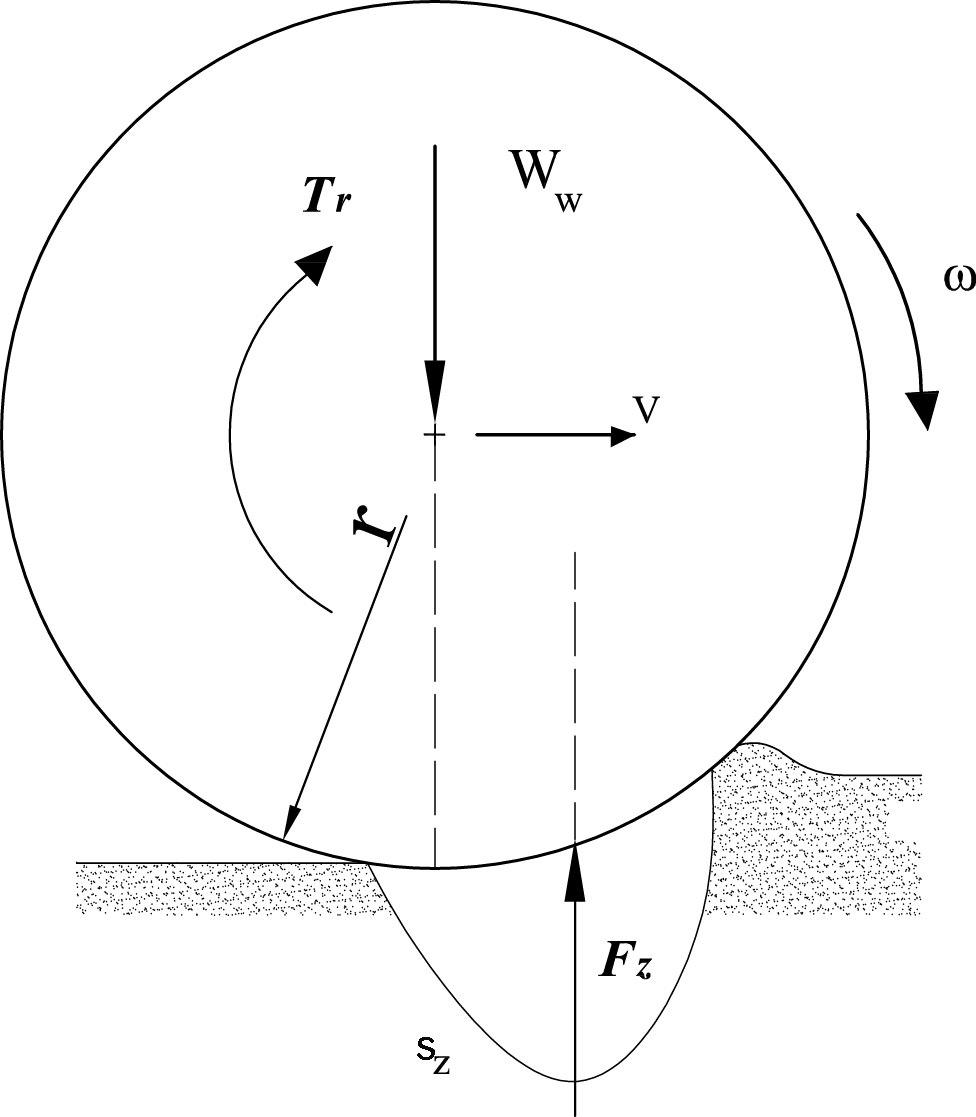}
      \caption{Motion resistance due to wheel-terrain interaction}
      \label{f5}
\end{figure}

\begin{figure}[t]
      \centering
      \includegraphics[width=8.6 cm]{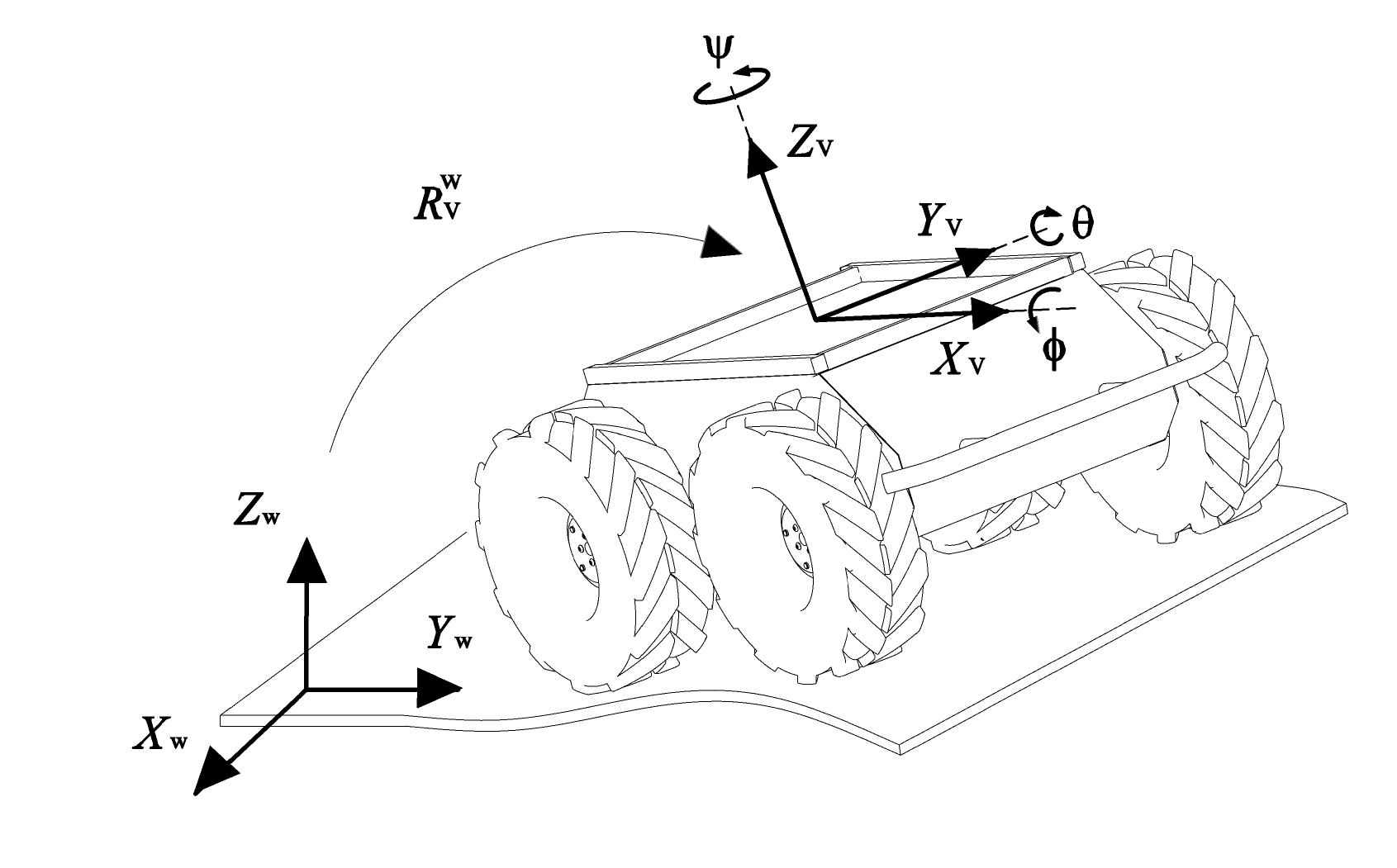}
      \caption{Reference frames and Euler angles for an all-terrain rover}
      \label{f6}
\end{figure}

\begin{figure}[t]
      \centering
      \includegraphics[width=6 cm]{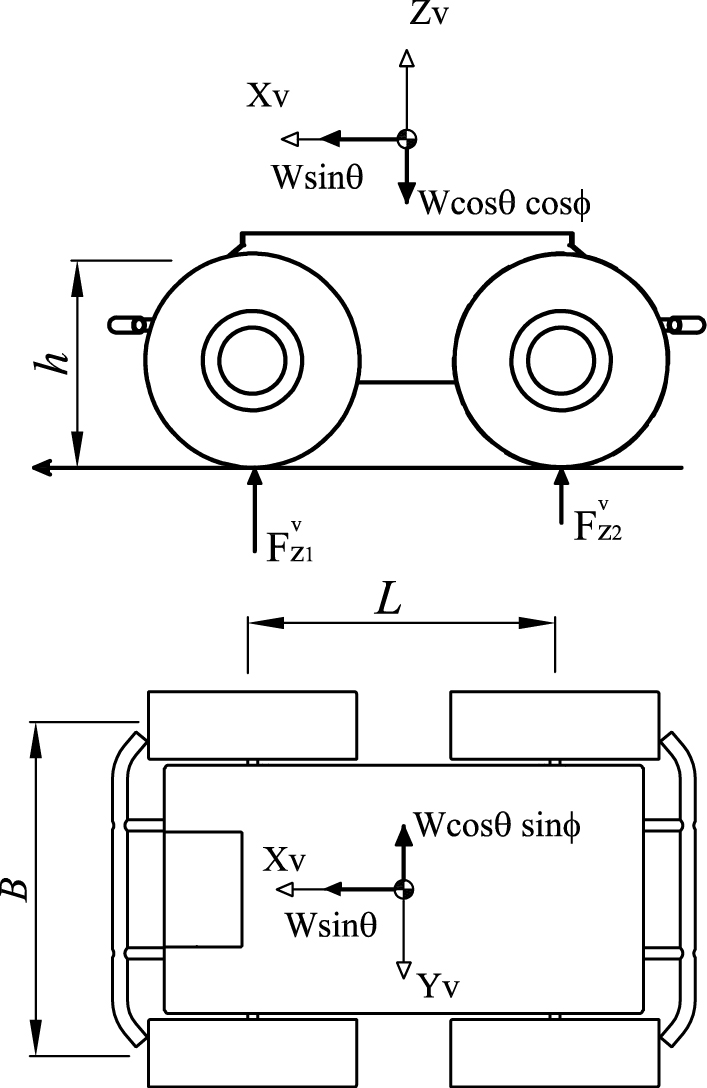}
      \caption{Vehicle load distribution for an arbitrary roll and pitch expressed in the vehicle reference frame}
      \label{f7}
\end{figure}

\begin{figure}[]
      \centering
      \includegraphics[width=8.5 cm]{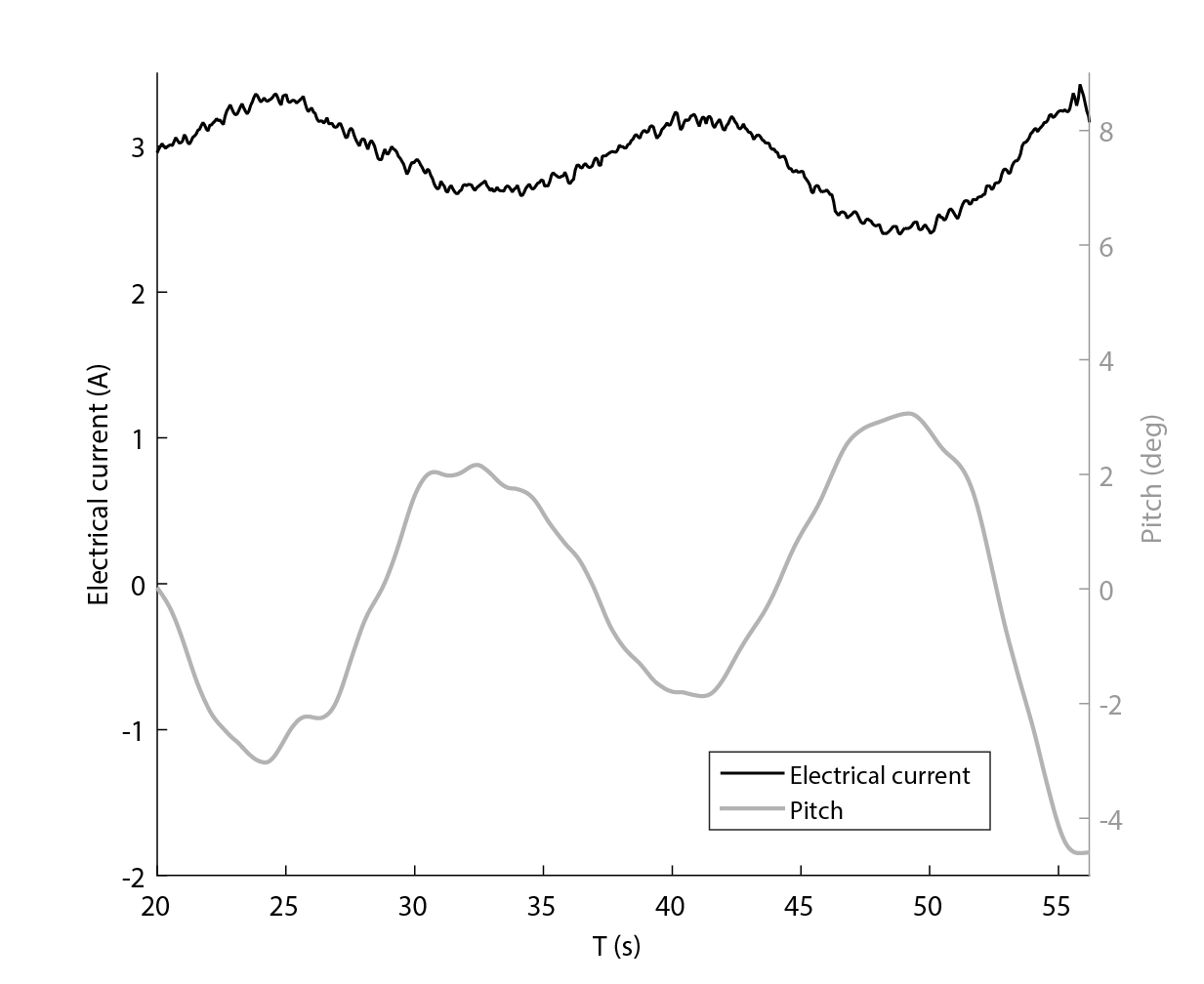}
      \caption{Operations on a sinuous terrain: tractive effort reflects the change in the vehicle pitch.}
      \label{f8}
\end{figure}

\begin{figure}[]
      \centering
      \includegraphics[width=8.5 cm]{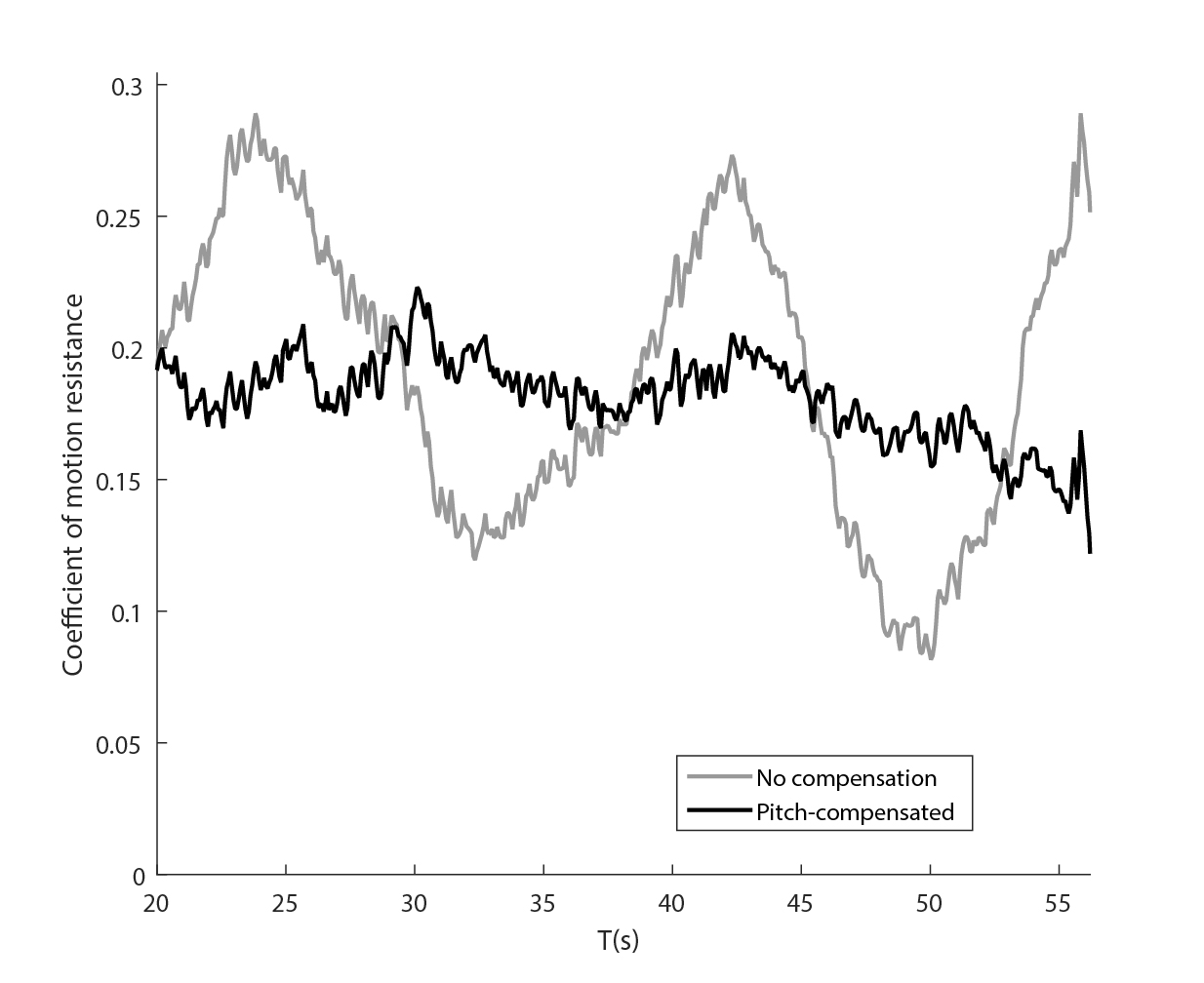}
      \caption{Estimation of motion resistance coefficient: accuracy improved when vehicle orientation was taken into account (black line), in contrast to no load distribution compensation (grey line).}
      \label{f9}
\end{figure}

On deformable terrain, a tyre with vertical load $W_v$ and linear and angular velocity, respectively, V and $\omega$, encounters a motion resistance that arises from the energy loss pertaining to the inelastic deformation of the bodies in contact. As a consequence, the resultant force of the normal stresses $s_z$ acting across the contact area, $F_z$, is shifted forward with respect to the wheel geometric centre, as shown in Figure \ref{f5}, requiring a certain amount of driving torque, $T_r$, to balance the corresponding resistance torque,
\begin{equation}\label{e12}
   T_r=f_r r F_\text{z}
\end{equation}
$f_r$ being the coefficient of motion resistance on a certain terrain, and $r$ the wheel radius. Tyre hysteresis due to the cyclic deformation of the contact patch is the prevalent of the motion resistance mechanisms on paved or hard terrain. Conversely, on deformable ground, the mechanical work for terrain compaction plays a dominant role, resulting in an increase of the motion resistance up to an order of magnitude \citep{BEK}.\\ The coefficient of motion resistance can serve as the first contact feature to discriminate a certain terrain. It can be indirectly estimated through the measurement of the electrical current $I$ that  is known to be roughly proportional to the mechanical torque $T_r$ in a DC brushed motor \citep{REI06, REI}
\begin{equation}\label{e13}
   T_r=\tau k_t I
\end{equation}
$k_t$ being the motor torque constant and $\tau$ the gearhead ratio (refer to Table \ref{t1} for more details). Therefore, the coefficient of motion resistance $f_r$ can be eventually expressed as a function of $I$ and $F_z$
\begin{equation}\label{e13}
   f_r=\frac{\tau k_t}{r}\frac{I}{F_z}
\end{equation}
One important aspect in the estimation of $f_r$ is the knowledge of the wheel vertical load. As a matter of fact, load distribution constantly changes during operations on irregular terrain. Since most of agricultural tasks are typically performed at constant and relatively low speed, a quasi-static model can be assumed to a first approximation neglecting the inertial contributions. Such a model can be solved for wheel vertical forces given the tilt of the vehicle. A world reference frame (WRF) $\{O_w, X_w, Y_w, Z_w\}$ and a vehicle reference frame (VRF) $\{O_v, X_v, Y_v, Z_v\}$ can be defined attached, respectively, to the ground and vehicle body, as shown in Figure \ref{f6}. The orientation of the vehicle with respect to the WRF can be defined by a set of Euler angles. The RPY convention was chosen in this research, that is, the sequence of rotations composed of roll ($\phi$), pitch ($\theta$), and yaw $(\psi$), respectively. The rotation matrix that defines the transformation from the VRF to the WRF is
\begin{equation}\label{e14}
    R^{W}_{V}=\left(
                  \begin{array}{c c c c c c}
                    c\psi\cdot c\theta && c\psi\cdot s\theta\cdot s\phi-s\psi\cdot c\phi && c\psi\cdot s\theta\cdot c\phi+s\psi\cdot s\phi\\
                    s\psi\cdot c\theta && s\psi\cdot s\theta\cdot s\phi+ c\psi\cdot c\phi && s\psi\cdot s\theta\cdot c\phi- c\psi\cdot s\phi\\
                    -s\theta && c\theta\cdot s\phi && c\theta\cdot c\phi\\
                  \end{array}
                \right)
\end{equation}
where $c$ and $s$, refer to $cos$ and $sin$, respectively.\\
If $F^W=[0, 0, -W]^T$ defines the weight force of the vehicle applied to its centre of mass and expressed in the WRF, the vehicle load distribution can be obtained by projecting $F^W$ in the VRF
\begin{equation}
F^{V}= \left( R^{W}_{V}\right)^{-1} F^W=\begin{bmatrix}
        W \sin{\theta}\\
        -W \cos\theta \sin\phi\\
       -W \cos\theta \cos\phi\\
    \end{bmatrix}
    \hspace{1mm}
    \label{eq_17}
\end{equation}
A free-body diagram of the system expressed in the VRF is shown in Figure \ref{f7}. By applying Newton's law and under the hypothesis that the external longitudinal and lateral forces are balanced proportionately by all four wheels, which is reasonable as the centre of gravity roughly coincides with the geometric centre of the vehicle, an analytical expression for the wheel vertical forces can be drawn in the VRF
\begin{equation}\label{e18}
   F_{z, 1}= \frac{W}{4}\cos\phi\cos\theta+\frac{W}{2}\sin\theta\frac{h}{L}-\frac{W}{2}\cos\theta\sin\phi\frac{h}{B}
\end{equation}
\begin{equation}\label{e19}
   F_{z, 2}= \frac{W}{4}\cos\phi\cos\theta-\frac{W}{2}\sin\theta\frac{h}{L}-\frac{W}{2}\cos\theta\sin\phi\frac{h}{B}
\end{equation}
\begin{equation}\label{e20}
   F_{z, 3}= \frac{W}{4}\cos\phi\cos\theta+\frac{W}{2}\sin\theta\frac{h}{L}+\frac{W}{2}\cos\theta\sin\phi\frac{h}{B}
\end{equation}
\begin{equation}\label{e21}
   F_{z, 4}= \frac{W}{4}\cos\phi\cos\theta-\frac{W}{2}\sin\theta\frac{h}{L}-\frac{W}{2}\cos\theta\sin\phi\frac{h}{B}
\end{equation}
$B$ being the vehicle width, $L$ its length, and $h$ the height of the centre of gravity above the ground.\\In conclusion, Eq.s (\ref{e18}-\ref{e21}) provide estimations of vehicle load distribution for an arbitrary roll and pitch angle. Given the vehicle orientation that can be estimated, for example, using an onboard inertial measurement unit and knowing the weight and geometric properties, the input to the problem, $F^V$ is defined, whereas outputs are the wheel vertical forces.
\\
As an example, Figure \ref{f8} shows the results obtained from the Husky vehicle during a straight drive at a constant speed of about 0.5 m s$^{-1}$ on an undulating terrain within an olive groove. The electrical current drawn by the drive motors is denoted by a black line whereas the vehicle pitch angle is plotted using a grey line. As the vehicle drives upward (downward),  the tractive  effort increases (decreases).
As a consequence, the vehicle load distribution changes. Figure \ref{f9} shows a comparison between the estimation in the coefficient of motion resistance with and without compensation for the change in the vehicle load distribution. The estimation accuracy improved when compensation was taken into account, that is using Eq.s (\ref{e18}-\ref{e21}). The corresponding black line in Figure \ref{f9} follows an approximately constant trend. In contrast, when no compensation was assumed, the estimation was biased with a definite shape.
\subsubsection{Vehicle slip}
\begin{figure}[t]
      \centering
      \includegraphics[width=14.0 cm]{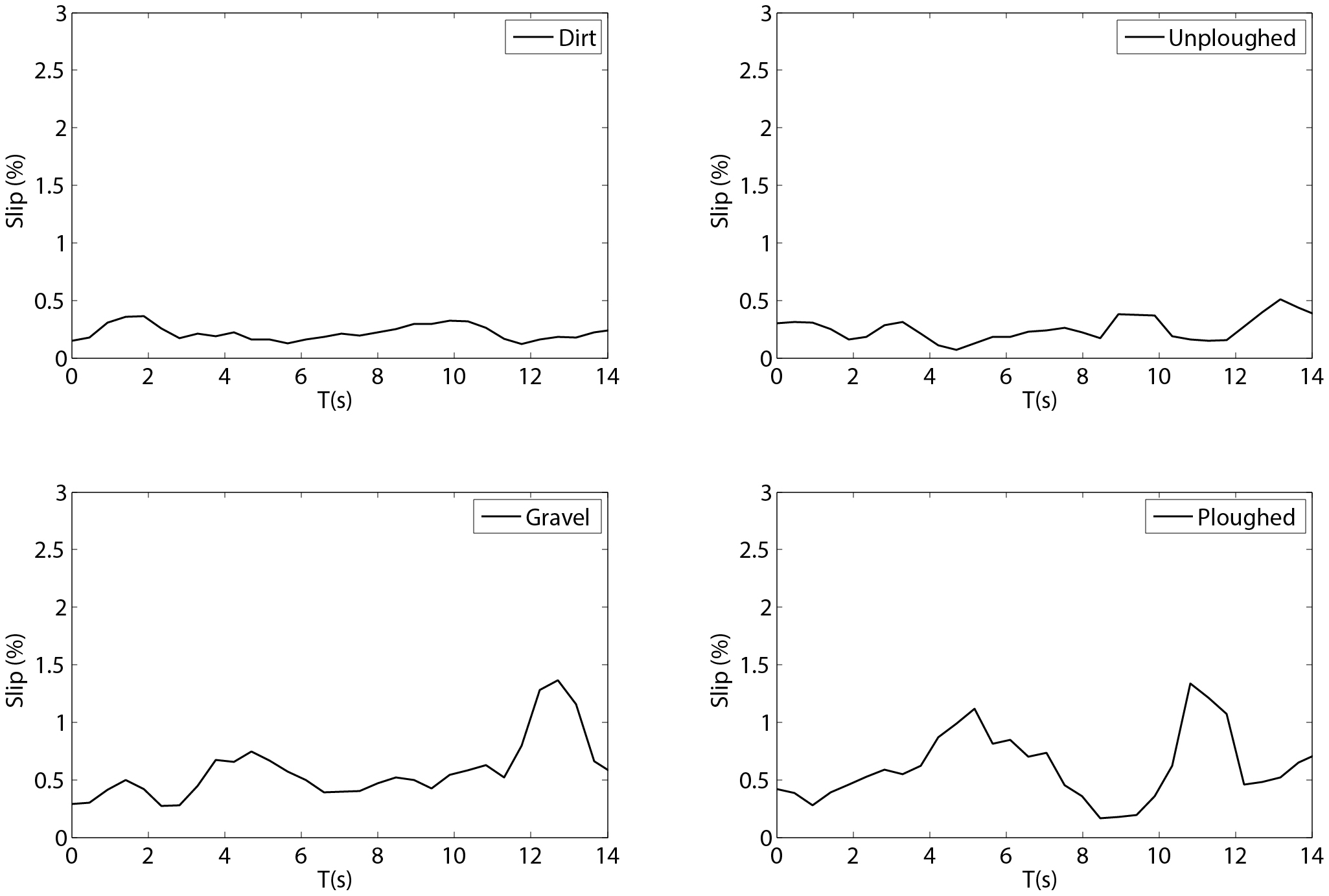}
      \caption{Examples of training slip data acquired during straight runs at constant speed of 0.5 m s$^{-1}$ on different terrains under zero-draught conditions}
      \label{fs}
\end{figure}

The second proprioceptive feature is the longitudinal slip incurred by the vehicle on a certain terrain. For the case of complaint tyre rolling on compliant ground, the ASABE Standard \citep{ASA} refers to this parameter also as the travel reduction. It is known that mobility on an unprepared surface is largely affected by the wheel-soil interaction. When a driving torque is applied to the wheel, a shearing action is initiated at the wheel-terrain interface generating a tractive effort used to overcome the rolling resistance and possibly accelerate the vehicle. As a consequence, the distance that the wheel travels is less than that in free rolling. This phenomenon is usually referred to as longitudinal slip. On certain terrains, a vehicle may experience a significant amount of slip and, in extreme cases, it can even get trapped due to a 100\% slip that can lead to mission failure, or can slow progress towards a desired goal. With reference to previous Figure \ref{f5}, slip $S$ can be defined as
\begin{equation}\label{e22}
   S=1- \frac{V}{V_t}=1-\frac{V}{\omega r}
\end{equation}
where $V$ is the actual forward speed of the vehicle and $V_t$ is its theoretical speed, which is equal to the product of angular speed $\omega$ and radius $r$ of the wheel. The undeformed radius calculated from the tyre size specification, i.e., $r$ = 0.165 m, was used as the radius in Eq. (\ref{e22}). The definition of wheel slip expressed by Eq. (\ref{e22}) can be extended to the entire vehicle during straight motion by averaging the encoder readings coming from all wheels. \\Therefore, travel reduction can be indirectly estimated by comparing the theoretical vehicle speed obtained from wheel encoders with the actual velocity using a vision-based localisation system. In this research, the open source visual odometry approach \emph{libviso2} \citep*{Geiger2011IV} was used to estimate the 6 DOF motion between subsequent frames, based on the minimisation of the reprojection error of sparse feature matches.\\ As an example, sample training data obtained from each of the terrain class of interest are shown in Figure \ref{fs}, during straight motion at constant velocity of 0.5 m s$^{-1}$ on relatively horizontal surface under zero-draught conditions. Differences in the slip experienced by the vehicle can be observed with increasing value moving from dirt road and unploughed terrain (average slip of about 0.21\% and 0.29\%) to gravel and ploughed terrain (average slip of 0.41\% and 0.57\%, respectively).

\subsubsection{Vibrational response}
\begin{figure}[]
      \centering
      \includegraphics[width=6 cm]{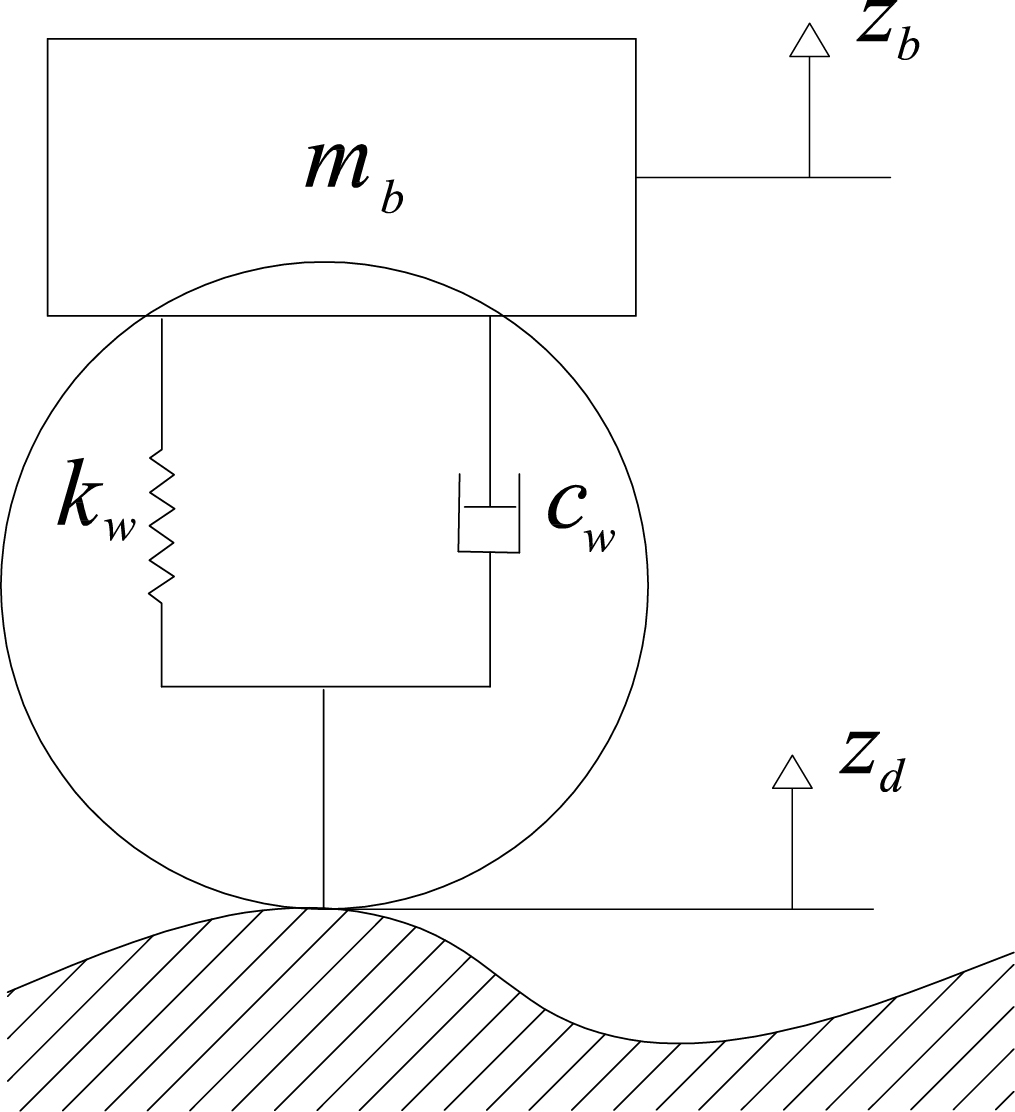}
      \caption{Quarter-vehicle model with one degree-of-freedom}
      \label{f10}
\end{figure}

\begin{figure}[]
      \centering
      \includegraphics[width=7 cm]{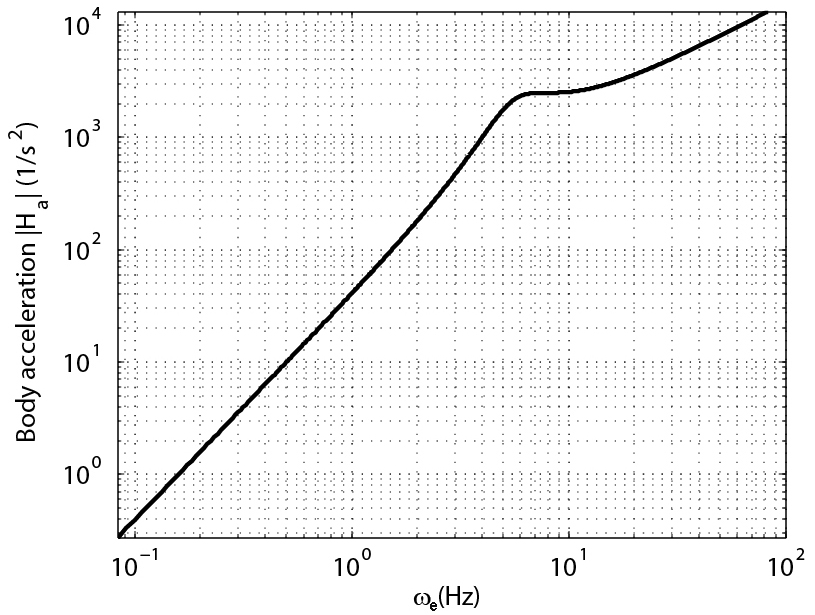}
      \caption{Frequency response of the Husky platform modelled as a quarter vehicle: ratio between the magnitude of the body acceleration and that of the ground displacement}
      \label{f11}
\end{figure}

\begin{figure}[!b]
      \centering
      \includegraphics[width=14.0 cm]{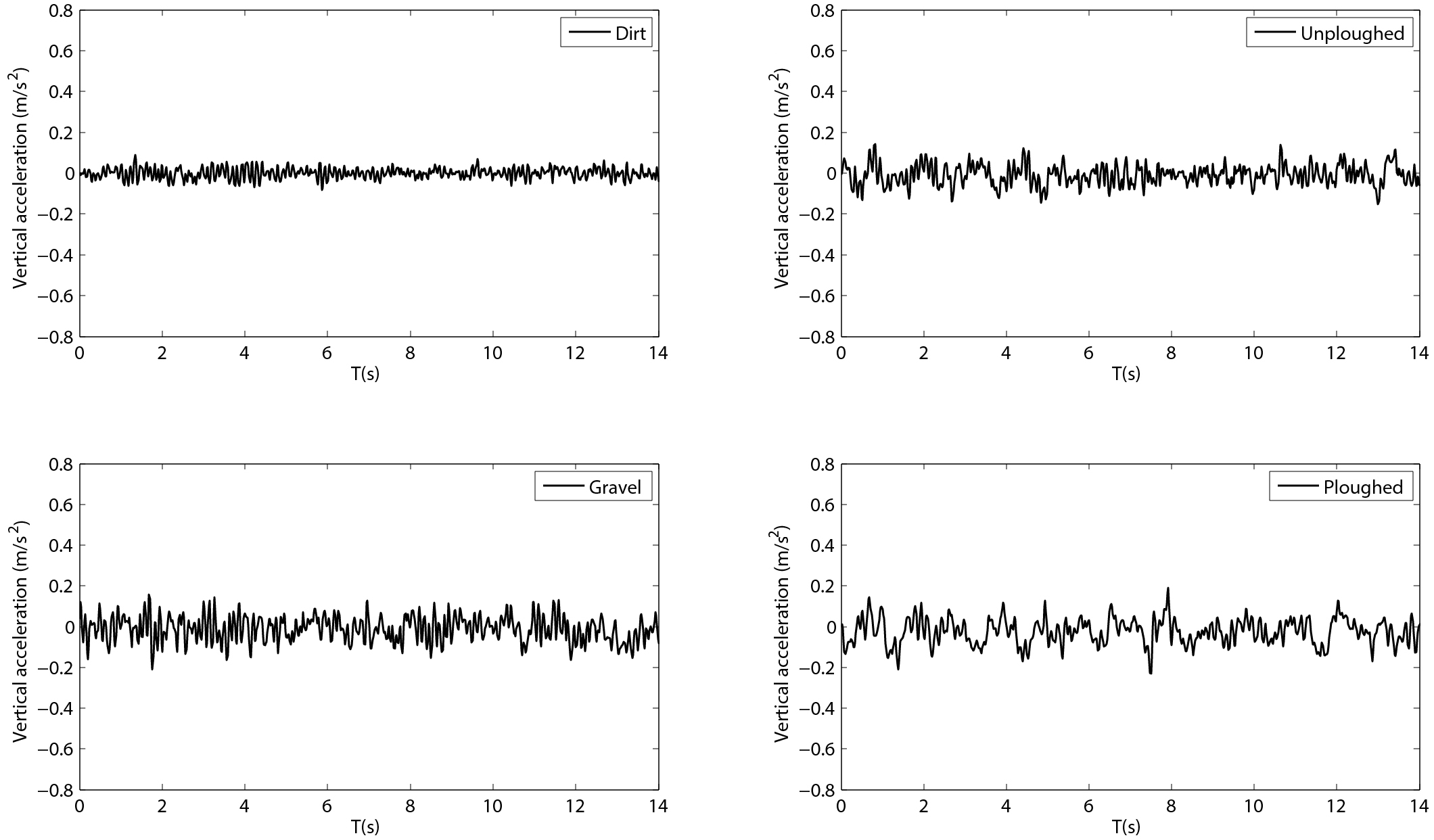}
      \caption{Acceleration data recorded during straight runs on the different terrains.}
      \label{f12}
\end{figure}

Analysis of vibrations propagating through a vehicles' structure can be used to distinguish between various types of terrain. In this research, the vertical acceleration of the vehicle sprung mass or body was used as the third distinctive feature for terrain characterisation. This measurement is obtained from the vertical accelerometer of the IMU that is mounted approximately in its centre of mass of the Husky platform.\\ The vibrations experienced by a vehicle due to terrain irregularities can be studied by referring to a quarter-vehicle (QV) model, as shown in Figure \ref{f10}. A one degree-of-freedom (1DOF) QV can be assumed that considers only the sprung mass ($m_b$) vertical displacement, $z_b$, since the Husky robot and most of agricultural tractors do not have a suspension system to restrict the body movement and preserve the accuracy of field operations. The vertical wheel stiffness and damping coefficient are, respectively, $k_w$ and $c_w$. The equation of motion for the QV model is given by
\begin{equation}
m_b\ddot{z}_b =k_w(z_d-z_b)+c_w(\dot{z}_d-\dot{z}_b)
\label{eq_qv}
\end{equation}
where $z_d$ is the vertical displacement of the terrain profile, that is the disturbance due to road irregularities. The relationship between the vertical response of the vehicle ($\ddot{z}_b$) and the harmonic excitation associated with a certain terrain ($z_d$) can be defined in the frequency domain by considering the magnitude of the transfer function for acceleration
\begin{equation}
|H_a(\omega_e)|=\frac{|\ddot{z}_b|}{|z_d|}=\omega_e^2\frac{|z_b|}{|z_d|}=\omega_e^2\sqrt{\frac{k_w^2+c_w^2\omega_e^2}{(k_w-m_b \omega_e^2)^2+c_w^2\omega_e^2}}
\label{eq_tf}
\end{equation}
$\omega_e$ being the excitation frequency defined as
\begin{equation}
\omega_e=2\pi\frac{V}{\lambda}
\label{eq_om}
\end{equation}
where $V$ is the travel speed and $\lambda$ is the irregularity wavelength. The frequency response of the Husky platform is plotted in Figure \ref{f11}, using the following realistic parameters in the model: $k_w$=10 kN m$^{-1}$, $c_w$=200 Ns m$^{-1}$, $m_b$=8 kg. If the vehicle travels at constant speed, the excitation mostly depends on the terrain profile. The smaller the wavelength, the higher the acceleration amplitude.
Typical terrain ``signatures" expressed in terms of accelerations along the $Z-$axis (compensated for the gravity component) are shown in Figure \ref{f12} during straight runs at an approximately constant speed of 0.5 m s$^{-1}$. It is observed that the signature of dirt road is very different from that of well-ploughed terrain with standard deviation of 0.026 m s$^{-2}$ and 0.084 m s$^{-2}$, respectively. Gravel and unploughed terrain show similar trends with standard deviation of 0.063 m s$^{-2}$ and 0.053 m s$^{-2}$, respectively.

In summary, the proprioceptive properties of a certain terrain patch were defined by the motion resistance coefficient, vehicle slip and vertical body acceleration. Considering the average value for motion resistance and slip, and the root mean square and standard deviation for acceleration, a combined four-element vector is drawn.
\section{Experiments}
This section reports the field validation of the proposed multi-modal classifier using the robotic platform described in Section \ref{s4}. The experimental setup is first described in detail. Then, experimental results are presented and the system performance are quantitatively evaluated in real agricultural settings.
\subsection{Experimental setup}
Figure \ref{ft} shows an aerial view  of the experimental farm located in San Cassiano, Lecce, Italy. It includes a vineyard and an olive grove that are connected by a dirt road. Four different types of terrain were considered that can be typically encountered in the farm and, in general, in the countryside of Salento:
\begin{figure}[t]
      \centering
      \includegraphics[width=10.0 cm]{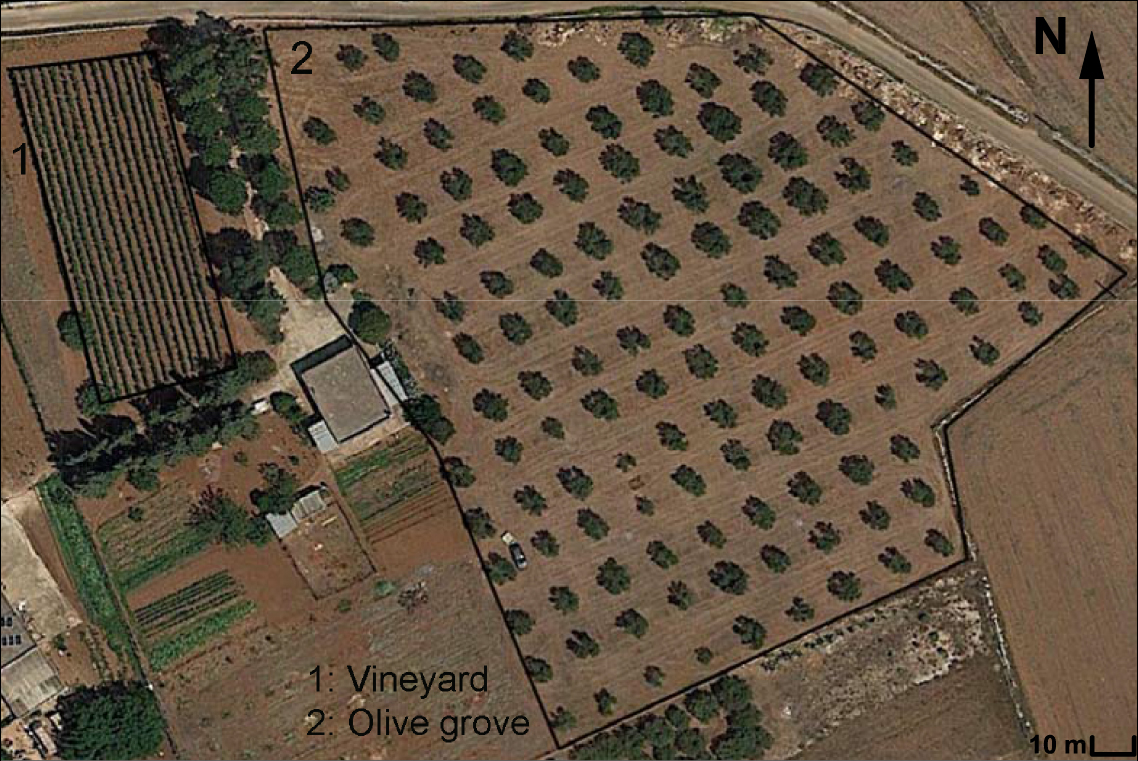}
      \caption{Experimental farm located in San Cassiano, Lecce, Italy. Aerial view taken from Google Earth (40$^\circ$03$^\prime$35.40$^{\prime\prime}$N, 18$^\circ$20$^\prime$50.98$^{\prime\prime}$E).}
      \label{ft}
\end{figure}
\begin{itemize}[\null]
  \item \emph{ploughed terrain}: mostly present in vineyard and consisting of farmland terrain broken and turned over with a plough. It is typically characterised by furrows and ridges;
  \item \emph{unploughed terrain}: unbroken agricultural land. It is a compact and relatively hard terrain and represents the vast majority of the terrain in the olive grove;
  \item \emph{dirt road}: unpaved road made of well-packed and compacted soil;
  \item \emph{gravel}: unconsolidated mixture of white/gray rock fragments or pebbles.
\end{itemize}
\begin{figure}[]
      \centering
      \subfigure[]{\includegraphics[width=4.2 cm]{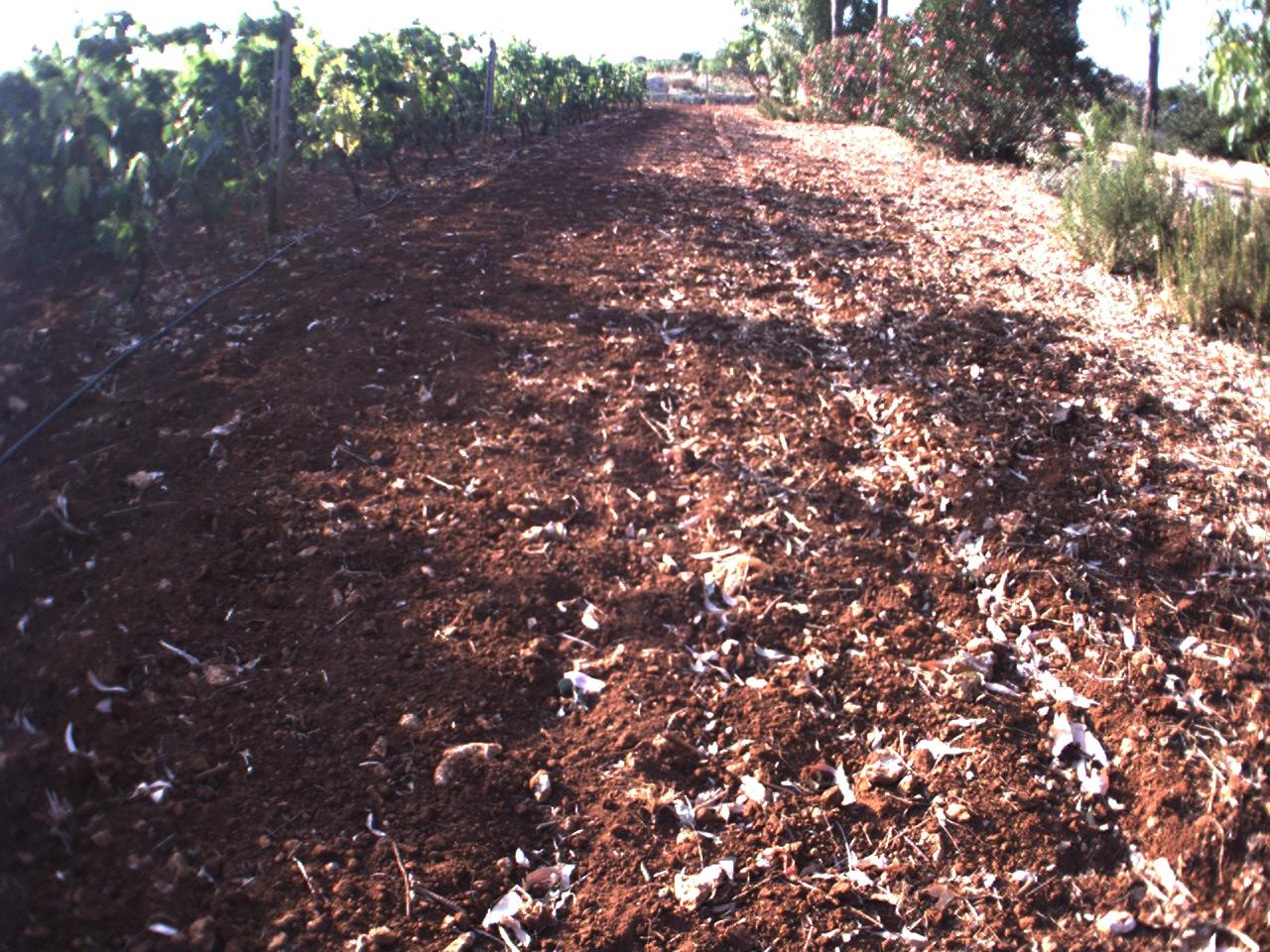}}
        \subfigure[] {\includegraphics[width=6.6 cm]{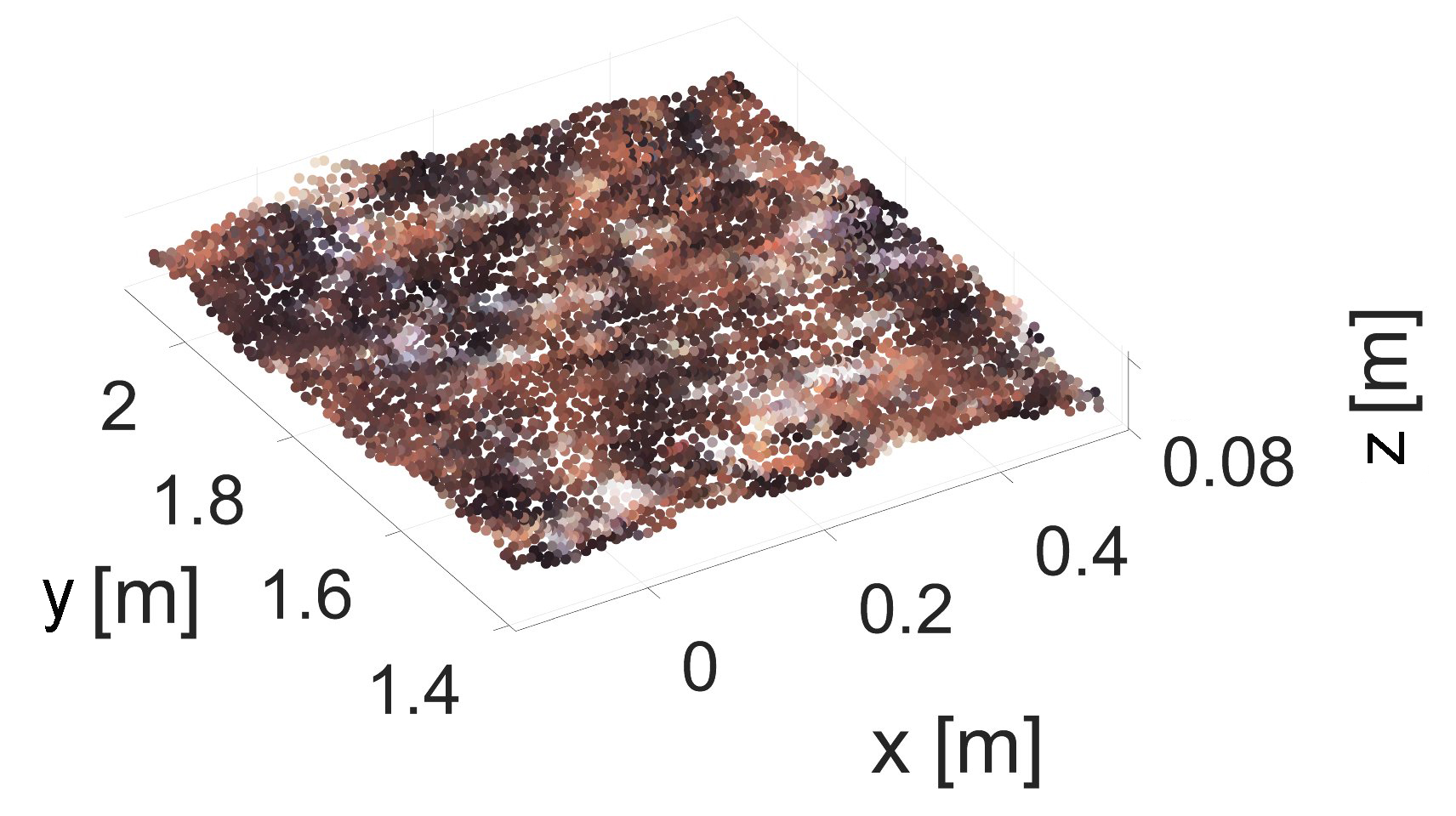}}\\
 \subfigure[]{\includegraphics[width=4.2 cm]{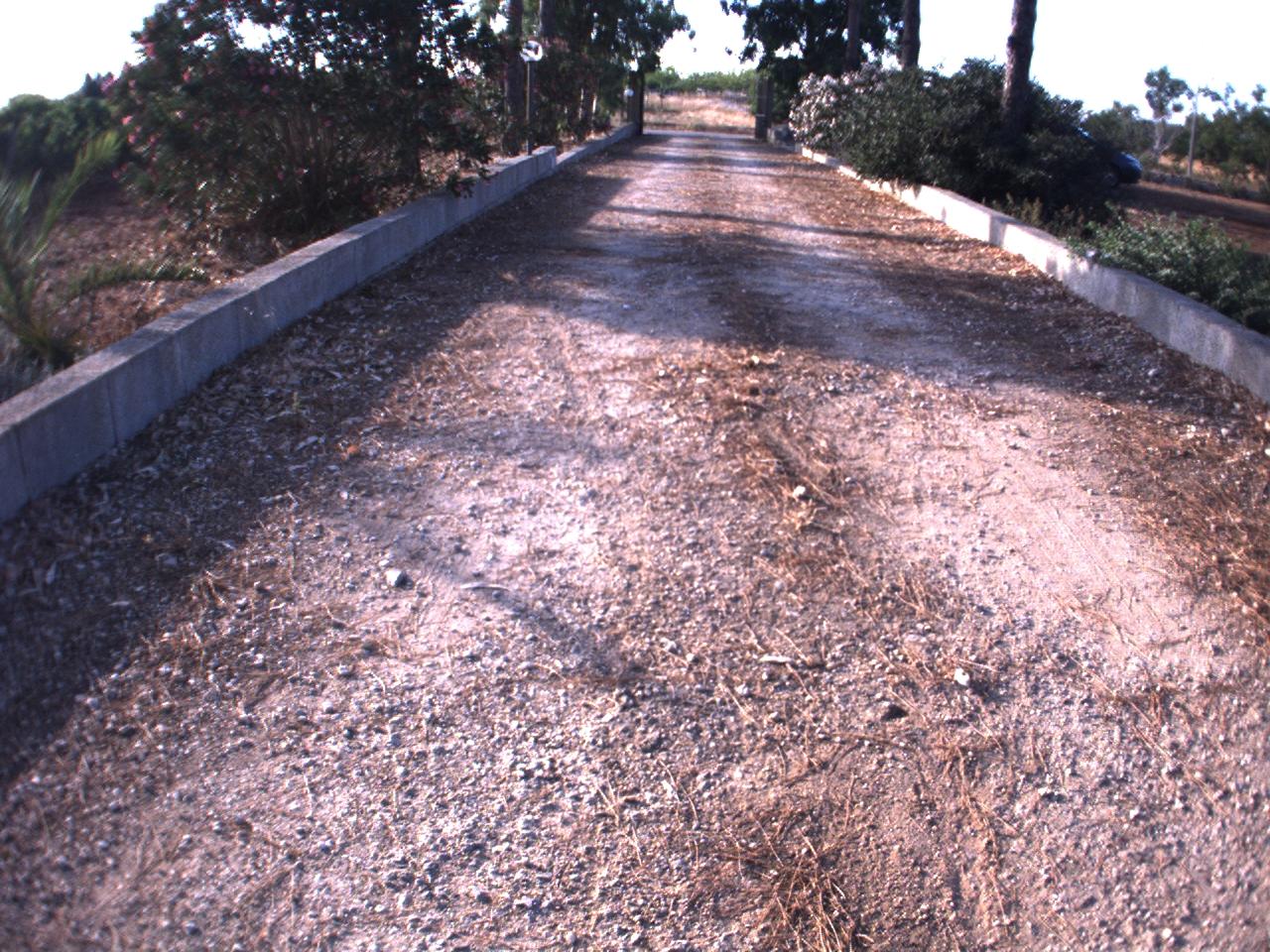}}
        \subfigure[] {\includegraphics[width=6.5 cm]{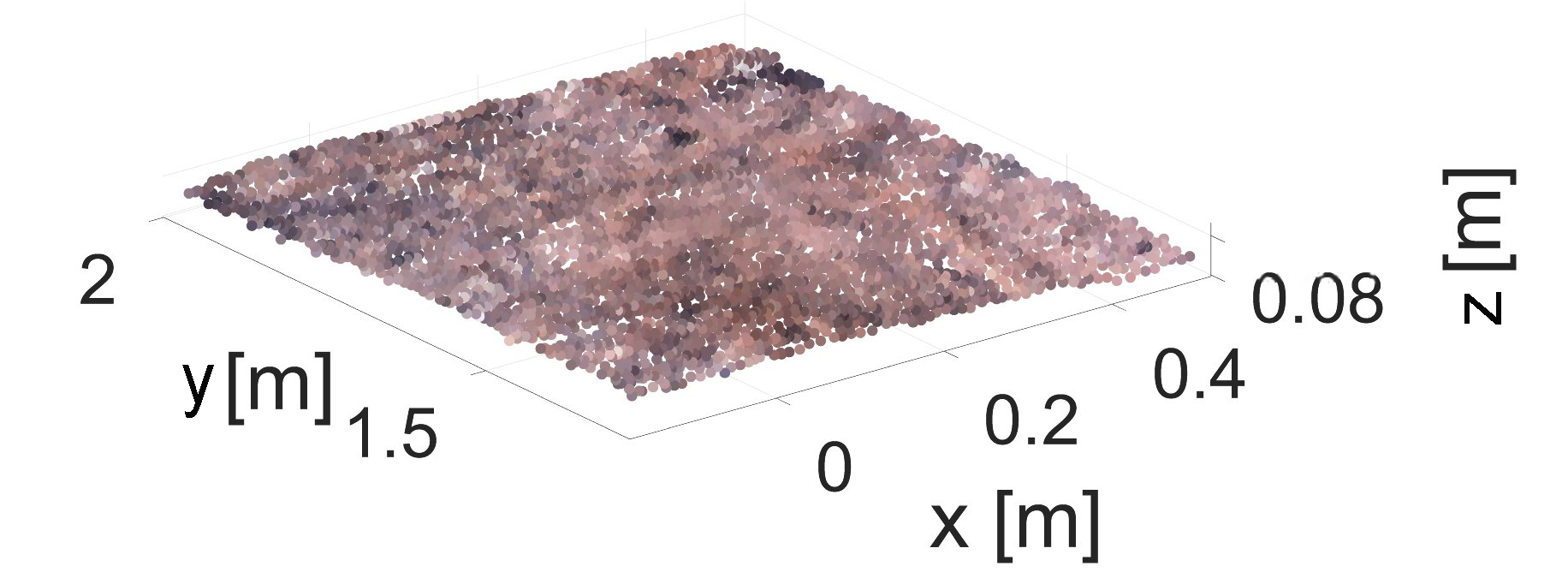}}\\
 \subfigure[]{\includegraphics[width=4.2 cm]{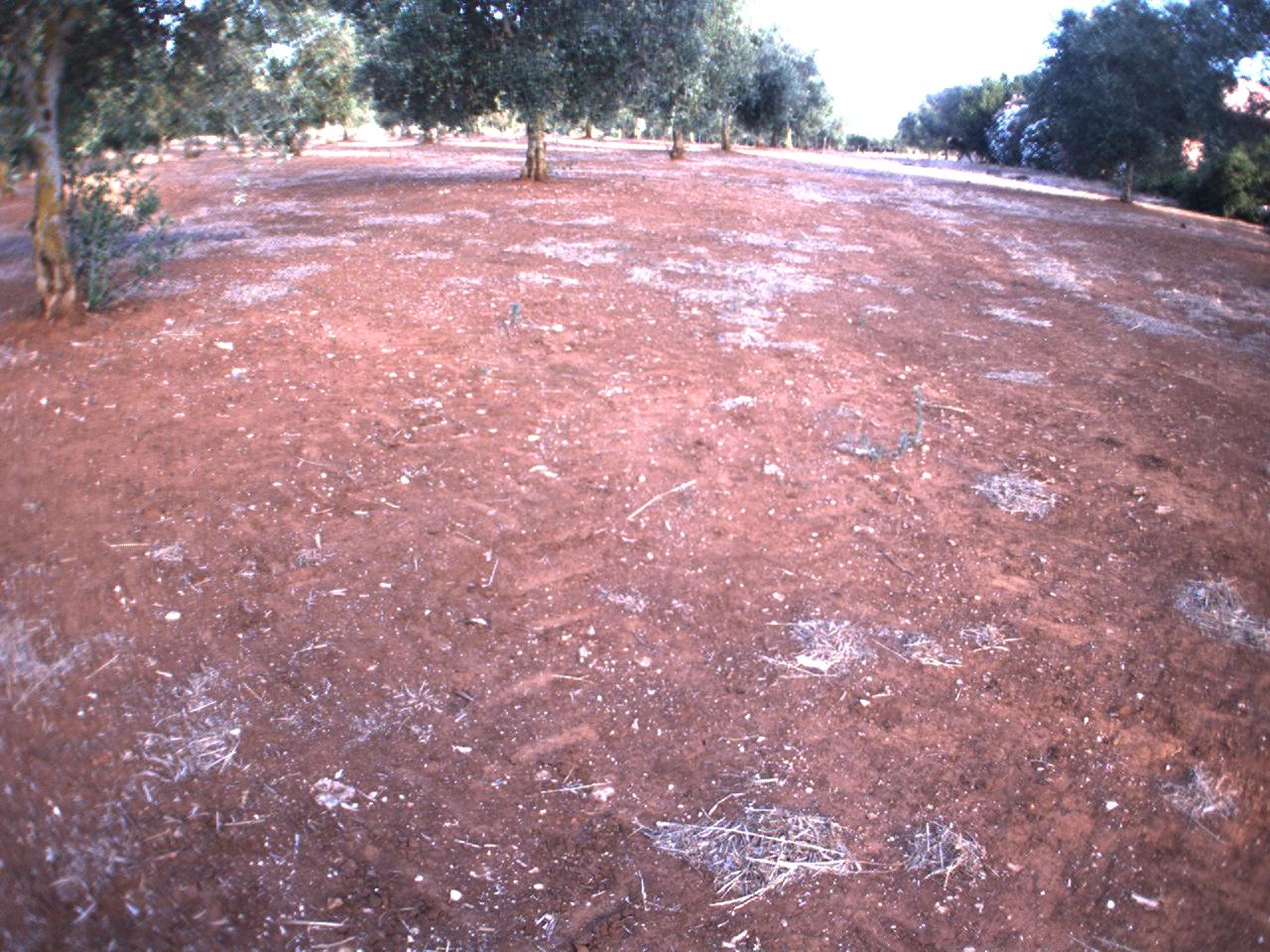}}
        \subfigure[] {\includegraphics[width=6.5 cm]{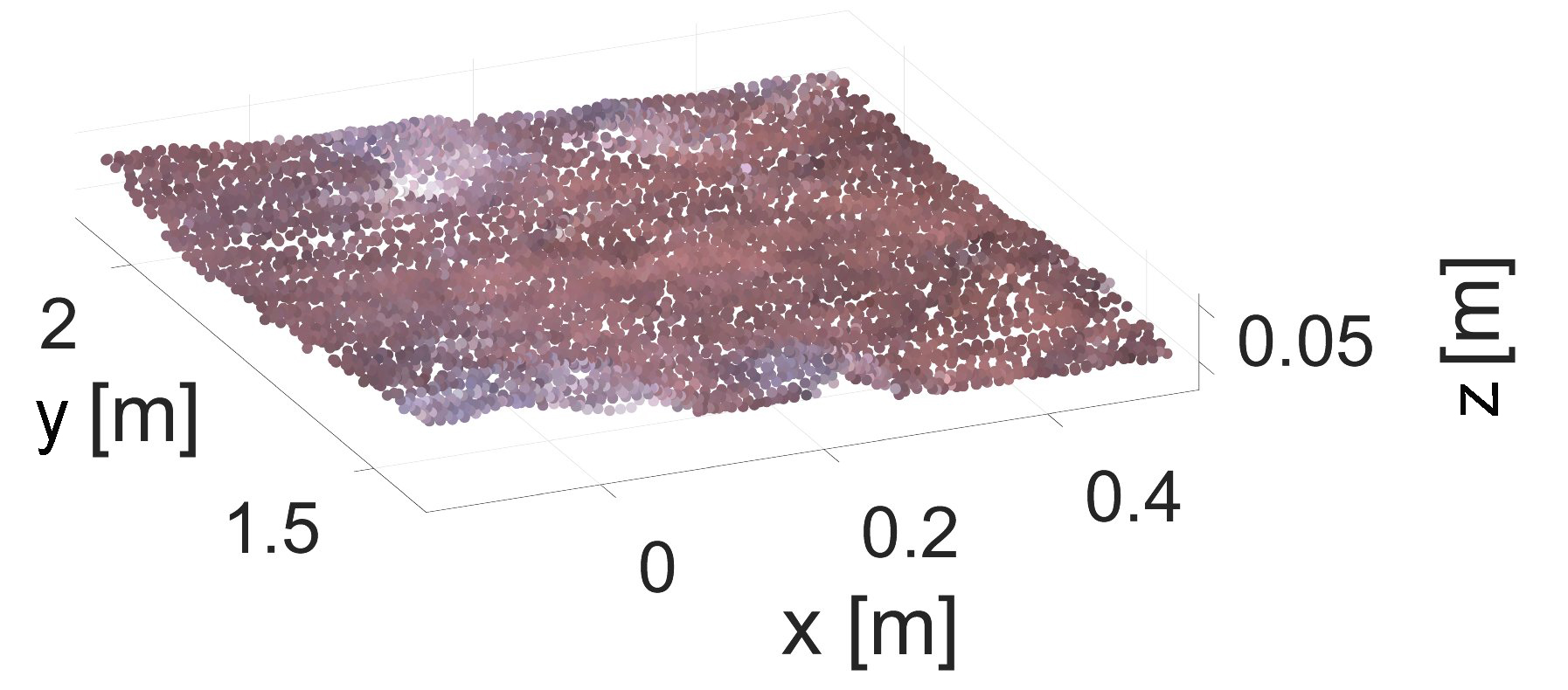}}\\
 \subfigure[]{\includegraphics[width=4.2 cm]{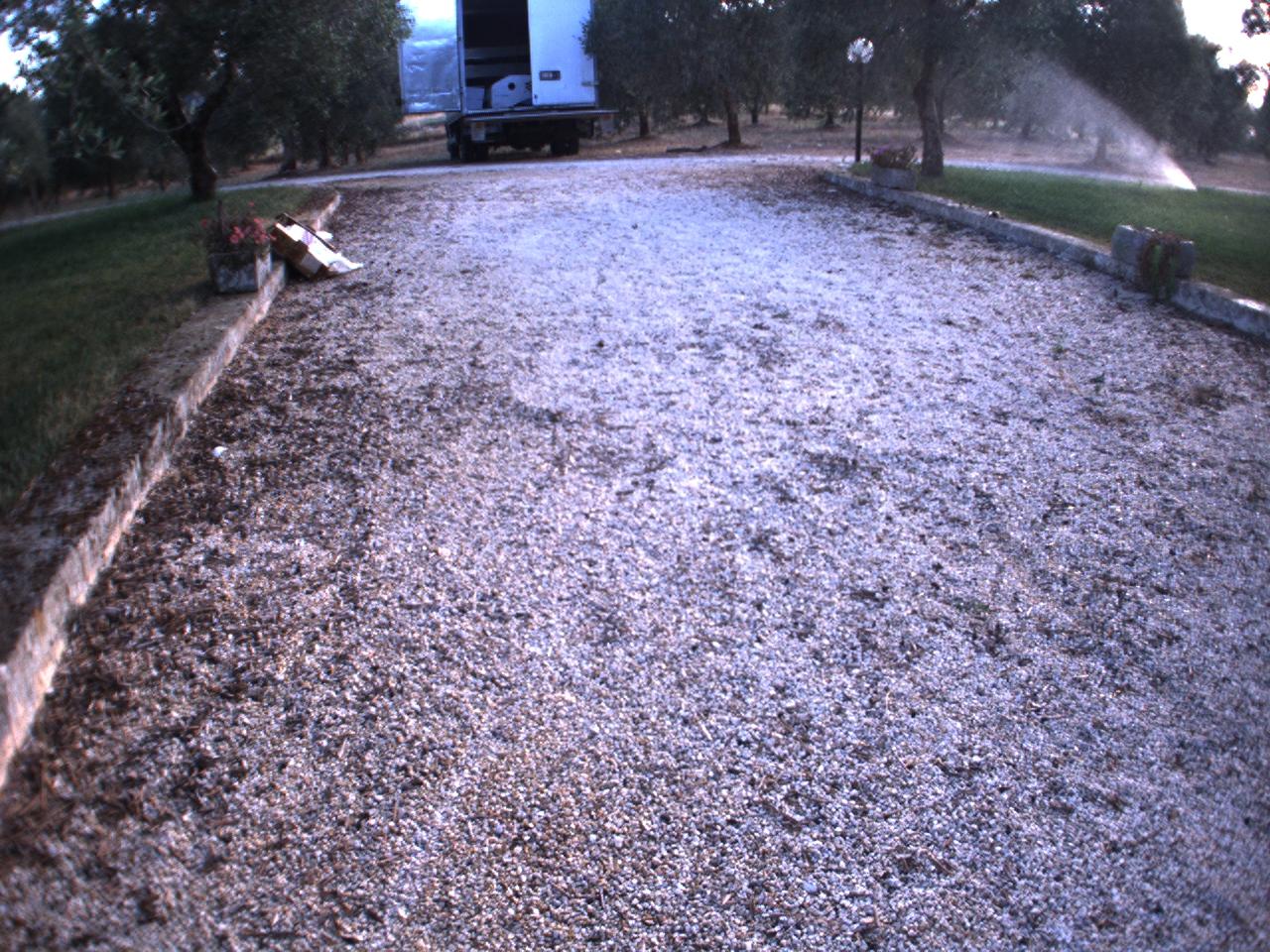}}
        \subfigure[] {\includegraphics[width=6.5 cm]{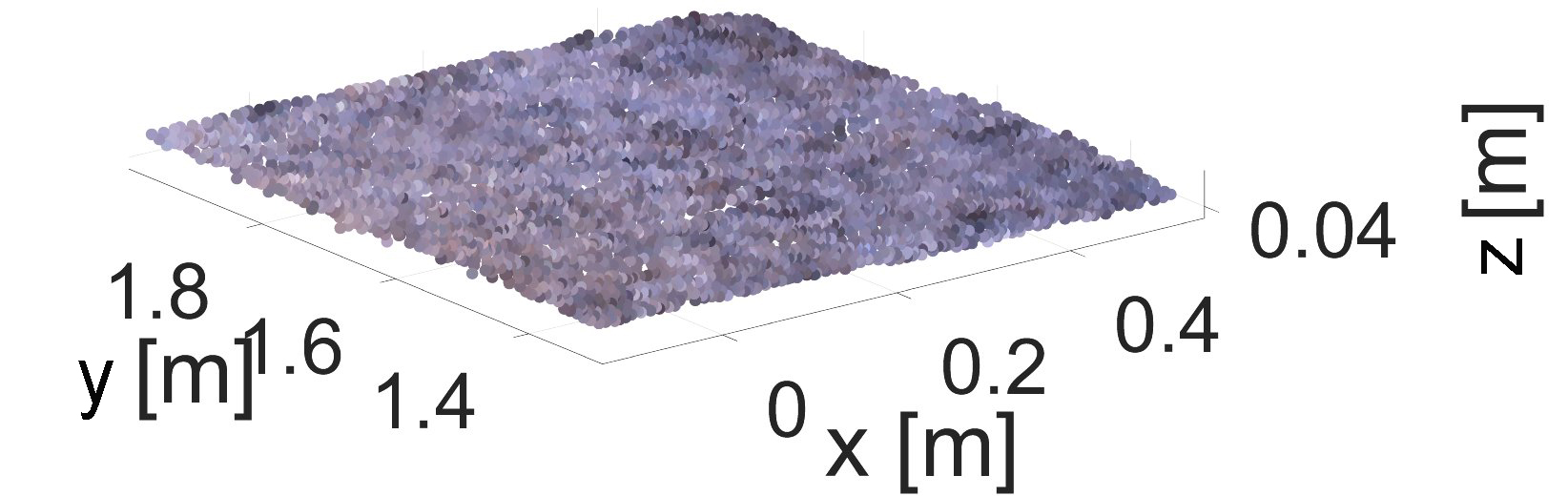}}\\
      \caption{Terrain classes:(a)-(b) ploughed terrain; (c)-(d) dirt road; (e)-(f) unploughed terrain; (g)-(h) gravel. Left: original sample images. Right: corresponding 3D terrain patches with RGB content obtained as output from the stereovision camera.}
      \label{classes}
\end{figure}
Sample images of the different terrain types are shown on the left side of Figure \ref{classes}. It can be seen that ploughed terrain is distinguishable from all other terrain classes in terms of colour, due to its relatively homogenous dark brown appearance, whereas dirt road and unploughed terrain both present a light brown colouring. In addition, dirt road includes parts with small rock fragments that may result in similarities with gravel, as well as desiccated foliage on the ground in unploughed terrain may lead to white/grey spots similar to gravel colour. Furrows and ridges in ploughed terrain are expected to result in distinctive geometric characteristics, while no significant differences in terms of geometric features can be expected among the other three terrain types.\\For each terrain class, an example of corresponding 3D patch as generated by the colour stereo-camera is shown on the right side of Figure \ref{classes}. In the context of this research, a terrain patch results from the segmentation of the stereo-generated three-dimensional reconstruction using a window-based approach. First, only the points that fall below the vehicle's body or undercarriage were retained, thus discarding parts of the environment not directly pertaining to the ground such as bushes, trees, and obstacles in general. Then, terrain patches are formed incorporating (i.e., stitching) data acquired in successive acquisitions using vision-based localisation. In our implementation, given the 8.5 Hz frame rate of the stereo device, an operating speed of 0.5 m s$^{-1}$, and the dimensions of the vehicle's chassis ($0.54 \times0.70$ m), four consecutive scans are used that correspond to a terrain patch size of approximately $0.70\times0.70$ m. When assuming a constant vehicle speed, it leads to terrain patches of approximately equal size, and therefore comparable with each other.\\
In all tests, the vehicle was controlled by an operator via a wireless joypad while sensory data were logged for successive off-line processing. The travel velocity was kept at an approximately constant value (0.5 m s$^{-1}$) to eliminate, on first approximation, the influence of velocity on the estimation of the proprioceptive features.
\subsection{Classifier performance evaluation}
\begin{figure}[]
      \centering
    \subfigure[] {\includegraphics[width=6.3 cm]{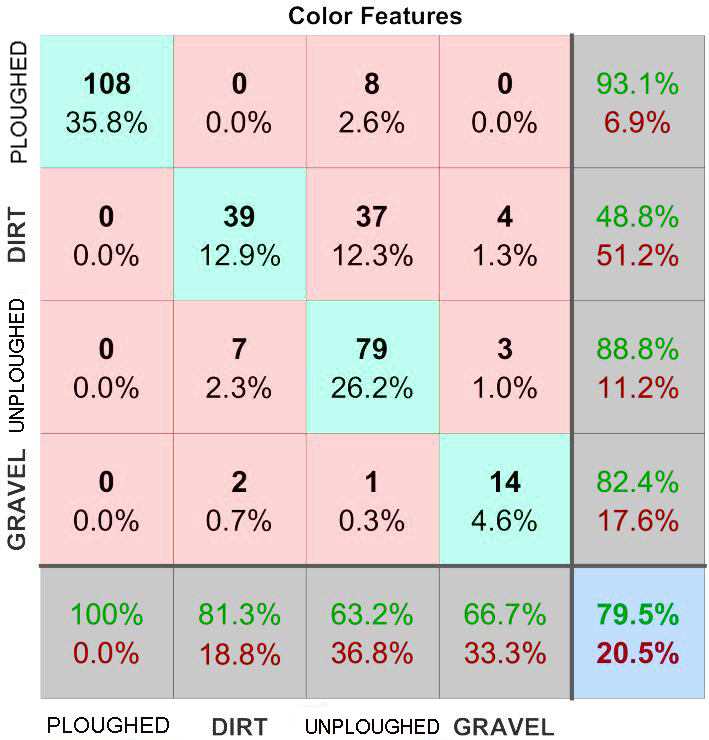}}
    \subfigure[] {\includegraphics[width=6.3 cm]{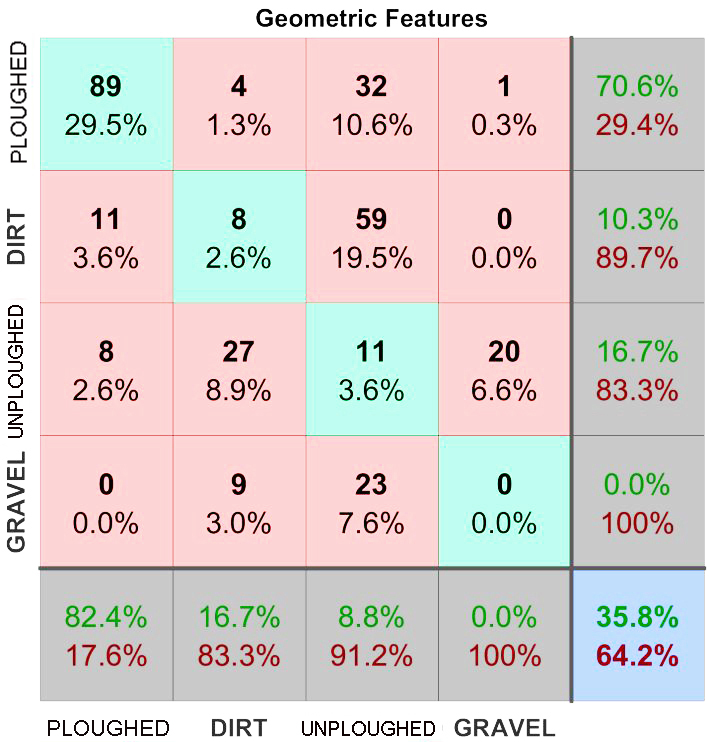}}\\
    \subfigure[] {\includegraphics[width=6.3 cm]{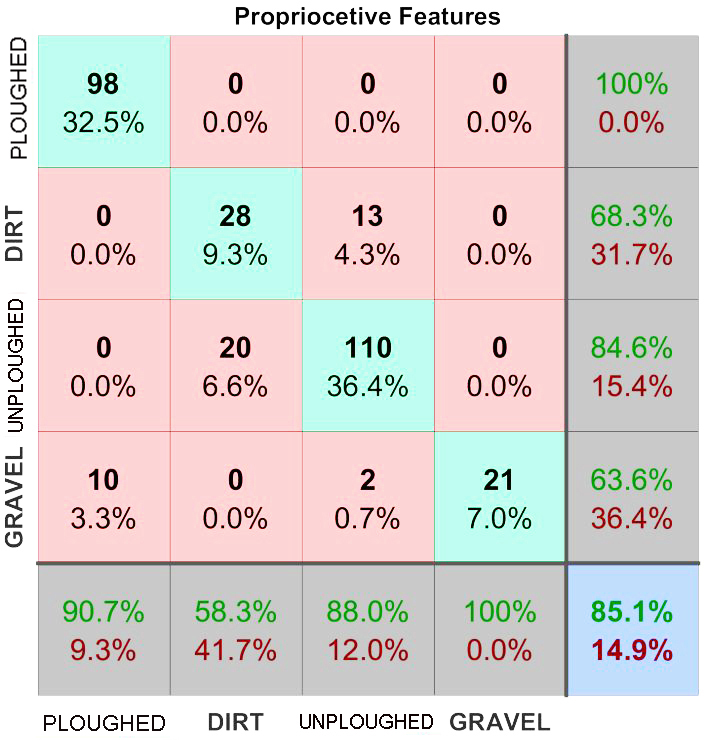}}
    \subfigure[] {\includegraphics[width=6.3 cm]{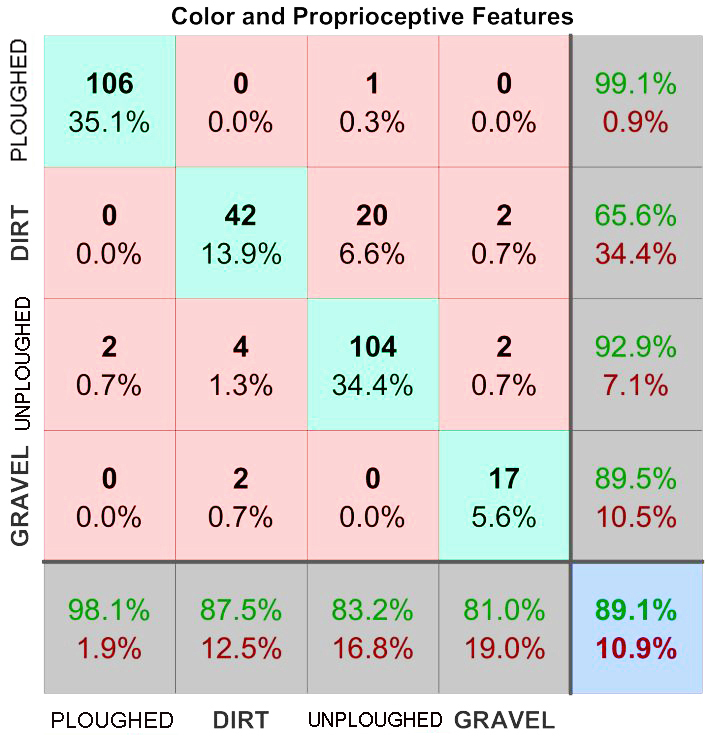}}
      \caption{Confusion matrices obtained for different feature sets: (a) colour; (b) geometric; (c) proprioceptive; (d) combination of colour and proprioceptive. Target and predicted classes are indicated along the horizontal and vertical direction, respectively}
      \label{confusion_matrices}
\end{figure}

Training data were gathered driving the vehicle on each terrain class along straight paths of at least 15 m. Independent tests are performed to acquire data for testing and evaluation of the classifier performance. Training and testing sets are randomly acquired at different hours of the day (i.e., from morning to mid afternoon) to account for lighting variations. \\For each class of interest, 59 samples are used for training.
A $k$-fold ($k = 5$) cross validation is performed to train the classifier. Specifically, the original sample is randomly partitioned into $k$ equally sized folds or divisions. A model is then trained for every fold using all the data outside the fold and is tested using the data inside the fold. Finally, the average test error over all folds is calculated.\\
First, it is discussed how the choice of the feature set impacts on the terrain classifier. Colour, geometric, and proprioceptive features are used singularly to train the classifier.\\ Once trained, the system was tested on an independent (i.e., different from that used for training) data set consisting of: 108 samples of ploughed agricultural terrain; 48 samples of dirt road; 125 samples of unploughed terrain; 21 samples of gravel. The unbalance of the test data results from the natural occurrence frequency of the given terrains in the investigated environment.  This suggests using per-class classifier performance measures in addition to average accuracy results \citep{POS}.
Specifically, in Tables \ref{tab:table1} to \ref{tab:table4}, recall (true positive rate), specificity (true negative rate), precision (positive predictive value), accuracy, and F1-score (i.e., the harmonic mean of precision and recall) estimates are shown for each terrain class.
\begin{table}[!b]
  \centering
  \caption{Classification results for the test data set using colour features}
  \label{tab:table1}
  \begin{tabular}{lccccc}
    Terrain type & Recall& Specificity& Precision& Accuracy& F1-score\\
    &(\%) & (\%) &(\%) & (\%) & (\%)\\
    \hline
    Ploughed terrain& 100.0 & 94.3 & 93.1 & 96.8 & 96.4\\
    Dirt road & 81.3 & 83.1 & 48.8 & 82.8 &60.9\\
    Unploughed terrain & 63.2 & 94.2 & 88.8 &81.1 &73.8\\
    Gravel& 66.7 & 98.7 & 82.4 &96.0 &73.7
  \end{tabular}
\end{table}
\begin{table}[]
  \centering
  \caption{Classification results for the test data set using geometric features}
  \label{tab:table2}
  \begin{tabular}{lccccc}
    Terrain type & Recall& Specificity& Precision& Accuracy& F1-score\\
    &(\%) & (\%) &(\%) & (\%) & (\%)\\
    \hline
    Ploughed terrain& 82.5 & 33.9 & 70.6 & 65.9 & 76.1\\
    Dirt road & 16.7 & 58.8 & 10.3 & 49.5 &12.7\\
    Unploughed terrain & 8.80 & 63.8 & 16.7 &39.0 &11.5\\
    Gravel& 0.0 & 77.1 & 0.0 &67.1 & $-$
  \end{tabular}
\end{table}
\begin{table}[]
  \centering
  \caption{Classification results for the test data set using proprioceptive features}
  \label{tab:table3}
  \begin{tabular}{lccccc}
    Terrain type & Recall& Specificity& Precision& Accuracy& F1-score\\
    &(\%) & (\%) &(\%) & (\%) & (\%)\\
    \hline
    Ploughed terrain& 90.7 & 100.0 & 100.0 & 96.3 & 95.1\\
    Dirt road & 58.3 & 94.6 & 68.3 & 88.6 & 62.9\\
    Unploughed terrain & 88.0 & 88.0 & 84.6 &88.0 &86.3\\
    Gravel& 100.0 & 95.1 & 63.6 &95.5 &77.8
  \end{tabular}
\end{table}
\begin{table}[]
  \centering
  \caption{Classification results for the test data set combining colour and proprioceptive features}
  \label{tab:table4}
  \begin{tabular}{lccccc}
    Terrain type & Recall& Specificity& Precision& Accuracy& F1-score\\
    &(\%) & (\%) &(\%) & (\%) & (\%)\\
    \hline
    Ploughed terrain& 98.1 & 99.4 & 99.1 & 98.9 & 98.6\\
    Dirt road & 87.5 & 91.2 & 65.6 & 90.6 & 75.0\\
    Unploughed terrain & 83.2 & 95.4 & 92.9 &90.3 &87.8\\
    Gravel& 81.0 & 99.2 & 89.5 &94.8 &85.0
  \end{tabular}
\end{table}\\
Corresponding confusion matrices are also shown in Figure \ref{confusion_matrices}. Target and predicted classes are indicated along the horizontal and vertical directions, respectively.  Diagonal cells show the number and percentage of correct classifications by the trained classifier, whereas the overall percentage of correct and wrong classifications are indicated in the bottom right cell.\\ The colour-based classifier (see Table \ref{tab:table1} and Figure \ref{confusion_matrices}(a)) performs relatively well with an accuracy always greater than 80\% and a best case of 96.8\% for ploughed terrain. F1-score ranges from 60.9\% for dirt road to 96.4\% for ploughed terrain. Misclassifications may arise especially due to colour similarities between different terrain classes. For instance, with reference to the corresponding confusion matrix, the proportion of correct classifications is of 63.2\% for unploughed terrain due to similar appearance with dirt road, resulting in low precision (i.e., 48.8\%) for the dirt road class. Conversely, ploughed terrain has a relatively homogeneous appearance that makes it clearly distinguishable from all other terrain classes, leading to high precision and recall values.
\\ The classifier trained upon proprioceptive features (see Table \ref{tab:table3} and Figure \ref{confusion_matrices}(c)) performs slightly better with an accuracy greater or equal to 88.0\% and F1-score ranging from 62.9\% for dirt road to 95.1\% for ploughed terrain.
\\
In contrast, geometric features (see Table \ref{tab:table2} and Figure \ref{confusion_matrices}(b)) seem to perform poorly in terms of accuracy (39.0\% and 49.5\% for unploughed terrain and dirt road, respectively) and F1-score (11.5\% and 12.7\% for unploughed terrain and dirt road, respectively). Overall, only 35.8\% classifications are correct, as shown by the corresponding confusion matrix. It should be added that the poor performance may be partly related to technological limitations of the stereo system. Higher-end modules would provide a better 3D reconstruction of the environment in terms of accuracy and resolution, possibly resulting in an improvement in the ability to classify different surfaces.\\
Based on these classification results, geometric features were discarded, whereas colour and proprioceptive features were retained and used in combination to train the terrain classifier. Feature combination was performed by concatenating colour and proprioceptive feature vectors. The two feature sets show complementary properties: for instance, for ploughed agricultural terrain and dirt road, colour features ensure higher recall and lower precision than the proprioceptive ones, while for unploughed terrain and gravel colour features exhibit lower recall and higher precision compared to the proprioceptive classifier. Therefore, their combination is expected to lead to overall better results. The specific adoption of proprioceptive features should help in differentiating between terrain types with similar colour (i.e., dirt road and unploughed terrain).\\Classification metrics obtained from the combined terrain classifier are  reported in Table \ref{tab:table4} and Figure \ref{confusion_matrices}(d). It resulted in an improvement in the classification of all terrain categories. The F1-score ranges from 75.0\% for dirt road to 98.6\% for ploughed terrain, with a total of 89.1\% of correct predictions outperforming both the colour-based (79.5\%) and the proprioceptive-based classifier (85.1\%).

\subsection{Multi-modal terrain mapping}
\begin{figure}[]
      \centering
    \subfigure[] {\includegraphics[width=13.0 cm]{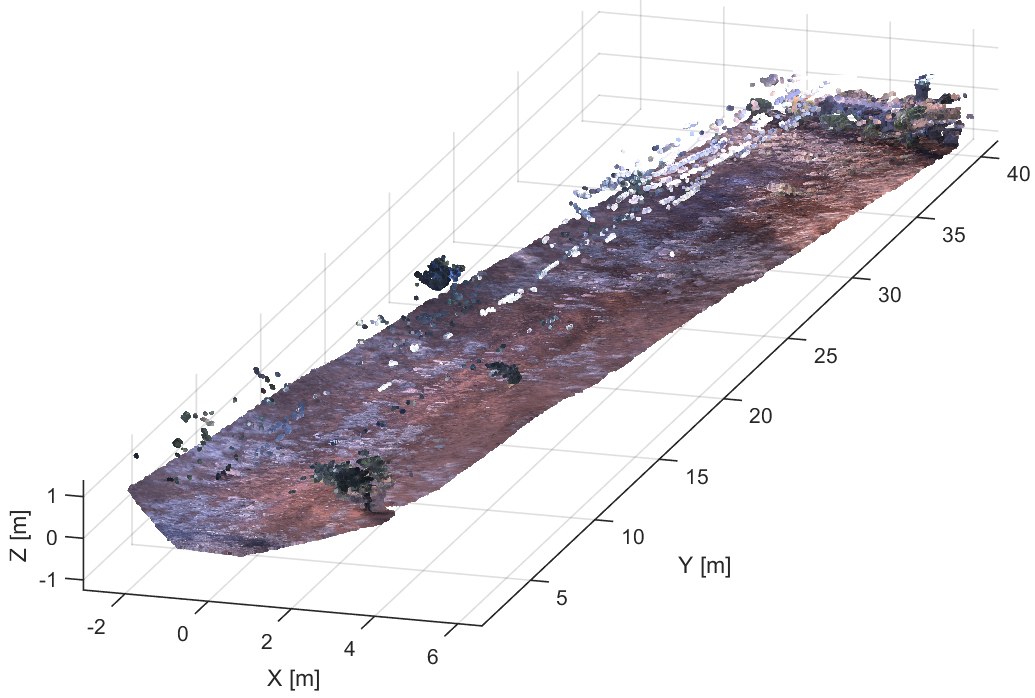}}
    \subfigure[] {\includegraphics[width=14.0 cm]{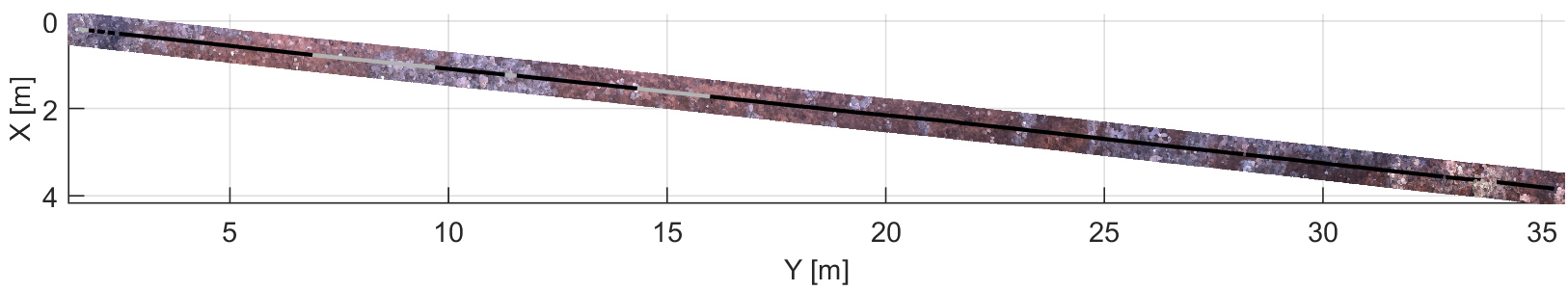}}\\
      \caption{Test on unploughed terrain in the olive grove. (a) 3D stereo map of the environment; (b) top view of the corresponding terrain map and vehicle path with overlaid classification results. Black solid line: unploughed terrain-labelled patch. Solid grey line: dirt road-labelled patch. Dotted black line: gravel-labelled patch.}
      \label{f15}
\end{figure}

\begin{figure}[]
      \centering
    \subfigure[] {\includegraphics[width=13.0 cm]{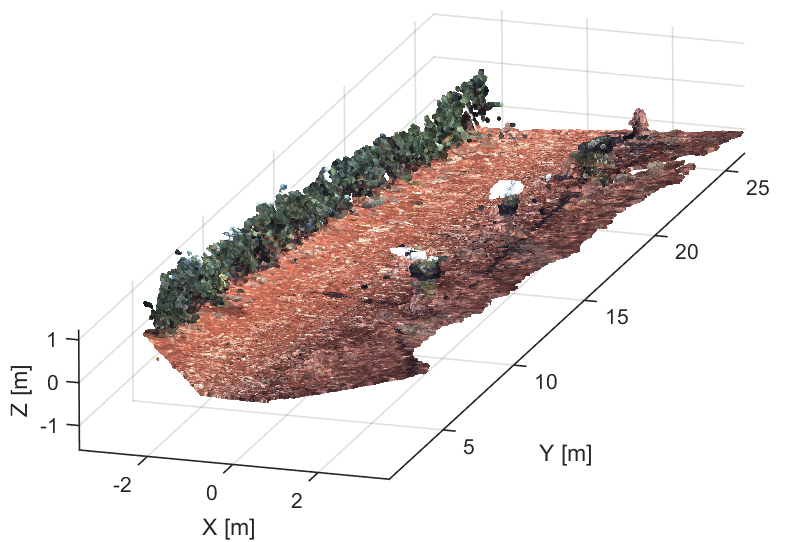}}
    \subfigure[] {\includegraphics[width=14.0 cm]{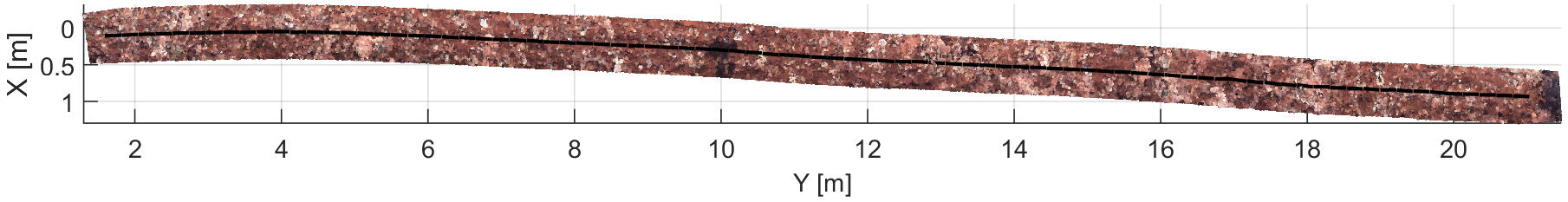}}\\
      \caption{Test on ploughed terrain in vineyard. (a) 3D stereo map of the environment; (b) Top view of the corresponding terrain map and vehicle path with overlaid classification results. Black solid line: ploughed terrain-labelled patch.}
      \label{test5_plowed_terrain}
\end{figure}

\begin{figure}[]
      \centering
    \subfigure[] {\includegraphics[width=13.0 cm]{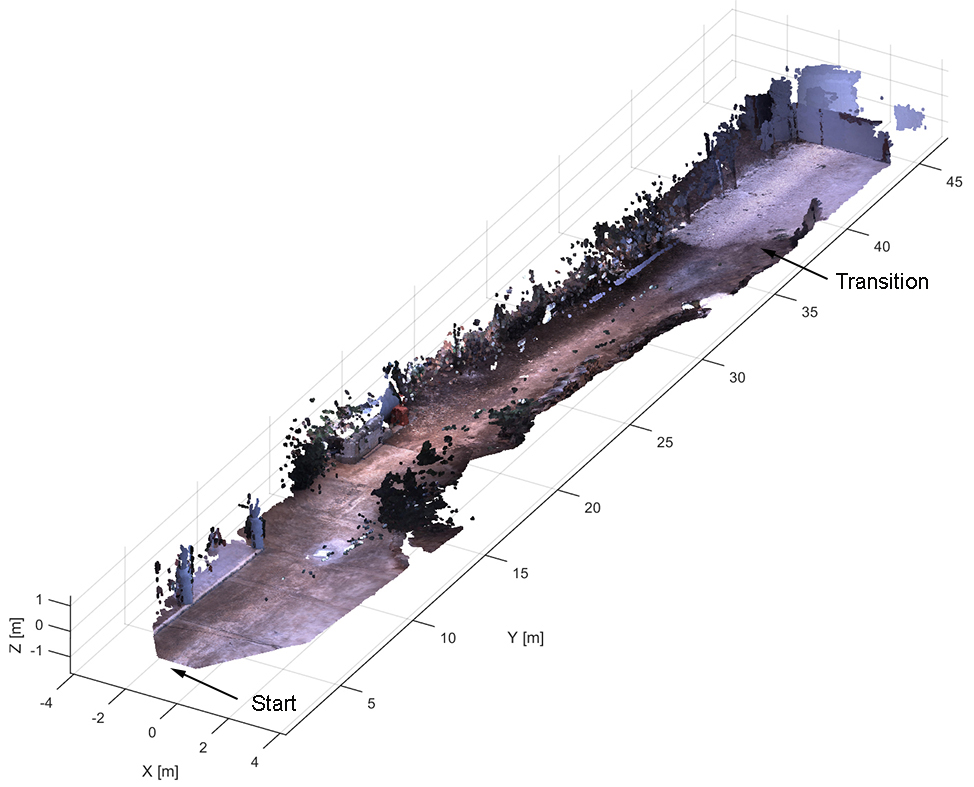}}
    \subfigure[] {\includegraphics[width=14.0 cm]{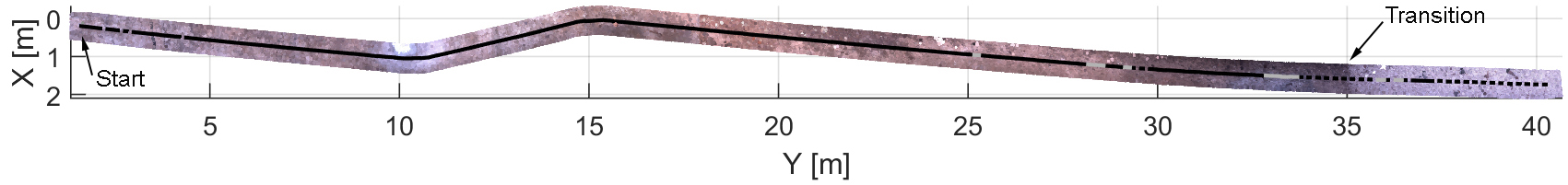}}\\
      \caption{Test on mixed terrain including dirt road followed by gravel. (a) 3D stereo map of the environment; (b) top view of the corresponding terrain map and vehicle path with overlaid classification results. Solid black line: dirt road-labelled patch. Dotted black line: gravel-labelled patch. Solid grey line: unploughed terrain-labelled patch.}
      \label{test1_mixed_terrain}
\end{figure}

\begin{figure}[t]
      \centering
    \includegraphics[width=14.2 cm]{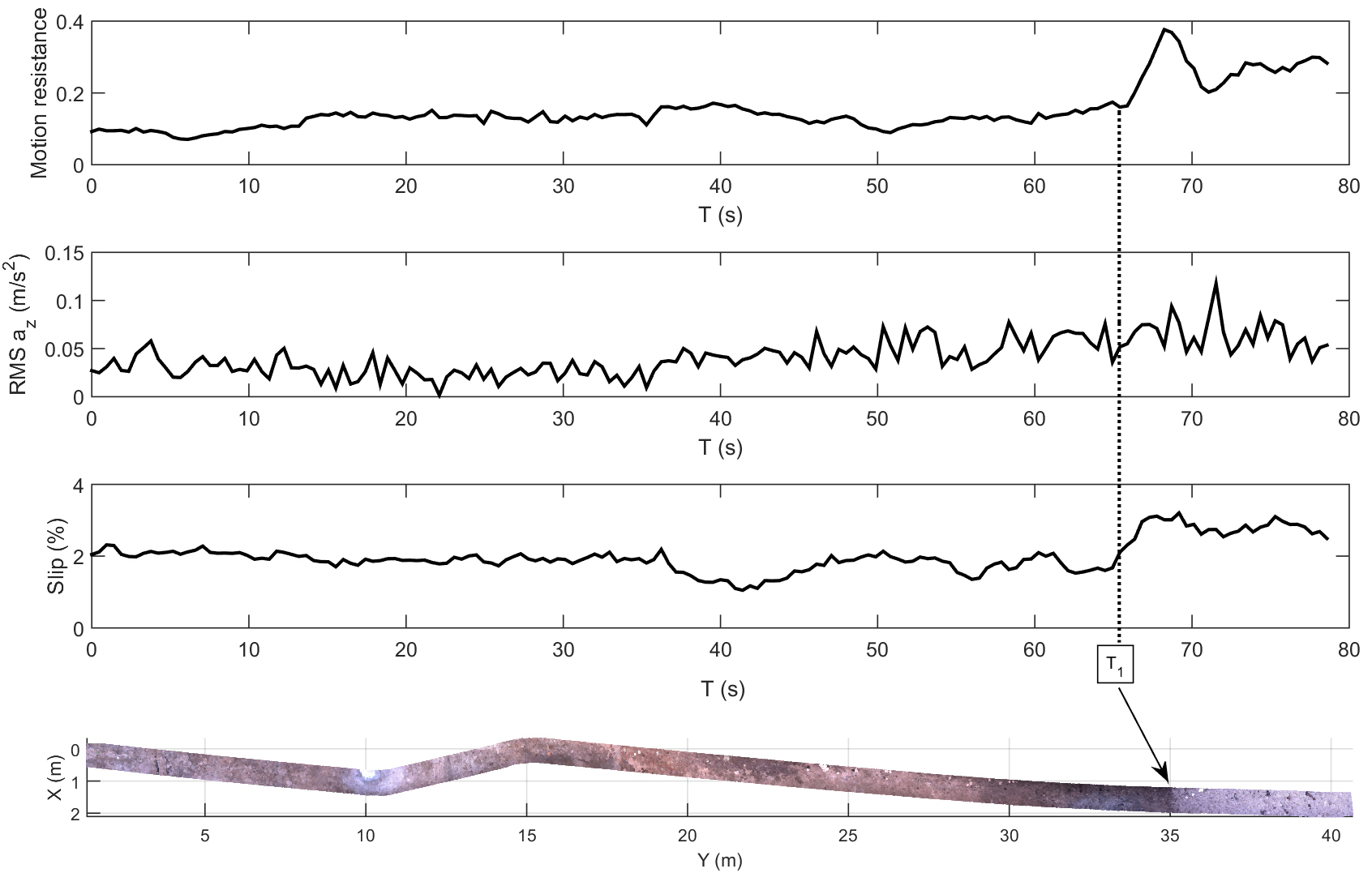}
      \caption{Multi-modal information for the test on mixed terrain (refer to Figure \ref{test1_mixed_terrain}). From top to bottom: motion resistance, RMS of vertical acceleration, slip and RGB-D information. At time $T_1$ the transition from dirt road to gravel is clearly visible in all graphs.}
      \label{multimodal_map}
\end{figure}

Underlying the proposed classification framework is the building of a multi-modal map of the traversed terrain, including 3D data, RGB content, and proprioceptive data (i.e., motion resistance, slip and vibrational response). This leads to a simultaneous multi-modal terrain mapping and classification approach, whereby the vehicle can build a map of the terrain and, at the same time, identify the category it belongs to. \\As an example, Figure \ref{f15} refers to a test where Husky was driven along an approximately straight path on unploughed terrain within the olive grove. The map of the environment reconstructed by the stereo system is shown in Figure \ref{f15}(a), while the corresponding terrain map with overlaid the path of the vehicle is shown in Figure \ref{f15}(b) as a top view. The combined classifier based on colour and proprioceptive features is applied to identify the traversed terrain. Results obtained from the classifier are shown along the path using different line types, that is, sections classified as unploughed terrain (correct classification) are marked by a solid black line, whereas dirt- and gravel-labelled sections (erroneous classification) are denoted by a solid grey and dotted line, respectively. The rate of correct predictions for this test is of 82.8\%.
\\
Figure \ref{test5_plowed_terrain} refers to another test where the vehicle traversed ploughed terrain along a vineyard row. Again, classification results are shown using different line types along the path. In this test, the classifier correctly identified all ploughed terrain observations (marked by a solid black line).\\Finally, Figure \ref{test1_mixed_terrain} shows the map with overlaid the classification results for a path on a mixed terrain. The vehicle started on dirt road and then moved to gravel. Terrain patches labelled by the system as dirt road are denoted by a solid black line, gravel-classified observations are marked by a dotted black line, and unploughed terrain is indicated by a solid grey line. The overall correct prediction rate for this test is of 85.5\%.
For completeness, Figure \ref{multimodal_map} visualises the multi-modal content of the terrain map including (from top to bottom) the coefficient of motion resistance, RMS of vertical acceleration, slip and RGB-D information. The transition from dirt road to gravel is clearly visible at time $T_1$.

\section{Conclusions}

This paper presented a framework for simultaneous mapping and classification of agricultural terrain. It uses common onboard sensors, that is, a vertical accelerometer and wheel encoders and torque sensors, and a colour stereo-camera. Using these components, the system can classify the terrain during normal vehicle operations.\\ The proposed classifier benefits from a multi-modal representation of the ground that combines exteroceptive (colour) with proprioceptive (motion resistance, slip, and acceleration) features within an SVM supervised approach. The two feature types show complementarity with the proprioceptive set helping in distinguishing between surfaces with similar colour. Validation of the system was performed in the field using an all-terrain vehicle operating in agricultural settings. Experimental results showed that the combined terrain classifier outperforms algorithms trained on single feature set with 89.1\% of correct predictions against the colour-based (79.5\%) and the proprioceptive-based classifier (85.1\%).\\This technology could be potentially used for all-terrain self-driving in agriculture. In addition, the generation of multi-modal terrain maps can be useful to inform farm management systems that would present to the user various data layers including RGB, 3D data, and resistance, slip and acceleration incurred by the vehicle.
\section*{Acknowledgment}
The financial support of the ERA-NET ICT-AGRI2 through the grant Simultaneous Safety and Surveying for Collaborative Agricultural Vehicles (S3-CAV) is gratefully acknowledged.

\section*{Author Contributions}
Giulio Reina and Annalisa Milella made significant contributions to the conception and design of the research. They mainly dealt with data analysis and interpretation, and writing of the manuscript. Rocco Galati focused on the experimental activity and data analysis (i.e., set-up of the system, design and realisation of experiments).

\bibliographystyle{harv_reina}
\bibliography{Bib}




\end{document}